\documentclass[5p,times]{elsarticle}
\usepackage{color}
\usepackage{lineno,hyperref}
\usepackage{verbatim}
\usepackage{graphicx}
\usepackage{subcaption}
\usepackage{mwe}
\usepackage{float}
\usepackage{graphics}
\usepackage{pdflscape}
\usepackage{xcolor}
\usepackage{amsmath}
\definecolor{brickred}{rgb}{0.8, 0.25, 0.33}

\usepackage[linesnumbered,ruled,vlined]{algorithm2e}
\usepackage{multirow}
\restylefloat{table}
\graphicspath{ {./images/} }

\journal{Journal of \LaTeX\ Templates}

\begin{document}

\begin{frontmatter}
\onecolumn\begin{center}
Thimal Kempitiya, Seppo Sierla, Daswin De Silva, Matti Yli-Ojanperä, Damminda Alahakoon, Valeriy Vyatkin: "An Artificial Intelligence framework for bidding optimization with uncertainty in multiple frequency reserve markets", Applied Energy, 280: 115918.
\\

https://doi.org/10.1016/j.apenergy.2020.115918
\\
© 2020 Elsevier. This accepted manuscript is made available under the CC-BY-NC-ND 4.0 license
\\
http://creativecommons.org/licenses/by-nc-nd/4.0/
\end{center}

\title{An Artificial Intelligence Framework for Bidding Optimization with Uncertainty in Multiple Frequency Reserve Markets}

\tnotetext[mytitlenote]{}


\author[latrobe]{Thimal Kempitiya\corref{cor1}}
\ead{t.kempitiya@latrobe.edu.au}
\author[aalto]{Seppo Sierla}
\author[latrobe]{Daswin De Silva}
\author[aalto]{Matti Yli-Ojanperä}
\author[latrobe]{Damminda Alahakoon}
\author[aalto,lulea]{Valeriy Vyatkin}

\cortext[cor1]{Corresponding author}

\address[latrobe]{Centre for Data Analytics and Cognition, La Trobe University,Victoria 3083, Australia }
\address[aalto]{Department of Electrical Engineering and Automation, School of Electrical Engineering, Aalto University, FI-00076 Espoo, Finland}
\address[lulea]{Department of Computer Science, Electrical and Space Engineering, Luleå Tekniska Universitet, SE-97187, Luleå, Sweden}

\begin{abstract}

The global ambitions of a carbon-neutral society necessitate a stable and robust smart grid that capitalises on frequency reserves of renewable energy. Frequency reserves are resources that adjust power production or consumption in real time to react to a power grid frequency deviation. Revenue generation motivates the availability of these resources for managing such deviations. However, limited research has been conducted on data-driven decisions and optimal bidding strategies for trading such capacities in multiple frequency reserves markets. We address this limitation by making the following research contributions. Firstly, a generalised model is designed based on an extensive study of critical characteristics of global frequency reserves markets. Secondly, three bidding strategies are proposed, based on this market model, to capitalise on price peaks in multi-stage markets. Two strategies are proposed for non-reschedulable loads, in which case the bidding strategy aims to select the market with the highest anticipated price, and the third bidding strategy focuses on rescheduling loads to hours on which highest reserve market prices are anticipated. The third research contribution is an Artificial Intelligence (AI) based bidding optimization framework that implements these three strategies, with novel uncertainty metrics that supplement data-driven price prediction. Finally, the framework is evaluated empirically using a case study of multiple frequency reserves markets in Finland. The results from this evaluation confirm the effectiveness of the proposed bidding strategies and the AI-based bidding optimization framework in terms of cumulative revenue generation, leading to an increased availability of frequency reserves.
\end{abstract} 

\begin{keyword}
Frequency Reserves, Smart Grid, Artificial Intelligence, Ancillary Markets, Bidding Strategies, Reschedulable Loads, Uncertainty Metrics, MC-Dropout, Artificial Neural Networks, Bayesian Neural Networks, Epistemic Uncertainty 
\end{keyword}

\end{frontmatter}


\section{Introduction}

 Power grid frequency deviations occur at any point in time when electricity production does not match consumption. Frequency reserves exist for the purpose of reacting to these deviations in real time, by modifying the electricity consumption or production behaviour of resources participating in the reserve. 
 Frequency reserves are an ancillary market with significantly different characteristics and behaviours compared to the primary day-ahead markets\cite{wang2011characteristics}. Frequency reserves markets are different from spot markets as the transaction is not the exchange of energy. Instead, it is the capacity to react to the grid under frequency situations by increased generation, reduced consumption or trading the frequency reserve capacity to the Transmission System Operator (TSO) \cite{giovanelli2018exploiting}.  For example, an electric vehicle (EV) charging station can participate on frequency reserves markets by curtailing the charging current during underfrequency. The TSO compensates the station operator for being in readiness to react to underfrequency situations. The compensation is proportional to the market price for that hour and the capacity (in kW) which the station is ready to curtail. Although numerous research endeavours have been conducted on spot market prediction and analysis  \cite{keles2016extended,saadaoui2017seasonal}, only a limited number have focused on these ancillary markets \cite{wang2013descriptive}. The need for new reserves is expected to increase due to the penetration of solar and wind power generation \cite{knap2015sizing} and the fact that reserves have traditionally been implemented with fossil fuel burning spinning reserves, which are now being eliminated in the push towards a carbon-neutral society \cite{delille2012dynamic}. Recent literature reports many innovative solutions for developing frequency reserve capability for distributed energy resources such as renewable power generation \cite{gholamrezaie2017optimal}, electricity storage \cite{shim2018harmonious}, electric vehicles \cite{moghadam2015distributed} and smart loads \cite{delavari2017improved}. However, in order to exploit such reserves in the most effective manner, it is necessary to trade these on frequency reserves markets. The statistical properties of frequency reserves markets are significantly different from spot markets, due to zero price days and exceptionally high price peaks\cite{giovanelli2018exploiting}. Thereby it is evident that several technical challenges need to be addressed in order to effectively trade frequency reserves capacities in multiple markets to generate value for all stakeholders.
 \\
 Firstly, the paper addresses the challenge of rescheduling the reserve capacity \cite{diaz2009scheduling}. Using electric vehicles as an example, since reserve provision is achieved by curtailing charging current, reserve market participation is only possible on market intervals in which the vehicle is plugged in and charging. In this case, rescheduling the charging is also rescheduling the market participation possibilities. For example, in Finland,  bidding on primary reserves markets is done day-ahead for each hour of the next day \cite{Oyj2018}, so the charging plans would need to be known at this time in order to profit from rescheduling. The market price often has large variations from one hour to the next \cite{giovanelli2018exploiting}. This implies that significant value can be generated by shifting frequency reserve capacity from low price hours to high price hours, based on quality market predictions available at the time of making the bid. 
 \\
 Secondly, many of the resources could be pre-qualified for more than one market, namely primary, secondary and tertiary frequency reserves. However, most research addresses only one market and a limited number of studies have focused on two markets \cite{zhang2018market,liu2018participation,ela2013market,zhang2014optimal}. An optimised bidding schedule for multiple markets depends on the national market rules, but week-ahead, day-ahead and hour-ahead market mechanisms are common. When frequency reserves have been pre-qualified for all accessible markets, the maximum value can be generated by selecting the market with the highest price forecast for each bidding period. 
 \\
 Finally, frequency reserves markets, unlike spot markets, are uniquely characterized by high price peaks and zero price days. It has been difficult for state-of-the-art forecasting methods to predict these irregularities accurately \cite{giovanelli2018exploiting}. This means bidding very high reserve capacities during forecasted price peaks and avoiding market participation during forecasted low prices may not be profitable if the bids were made based on uncertain price forecasts. 
 \\
 We report three research contributions in this article. The first contribution is the study of global frequency reserves markets that informs the design of a generalised model of the operation of a frequency reserves market. The second contribution is the formulation of three bidding strategies for multi-stage markets, based on this generalised model. The first two strategies are for non-reschedulable loads, while the third is for reschedulable loads with four alternative rescheduling schemes. These strategies are devised to account for market rules and data-driven predictions with uncertainty metrics. 
The third contribution is the use of Artificial Intelligence (AI) to forecast the multi-stage frequency reserve market prices, and to quantify the uncertainty of these forecasts, so that this information is available to the bidding strategies,
 followed by a demonstration of optimized revenue generation and predictive accuracy by empirically evaluating this AI-based bidding optimization framework on a real-world frequency reserves setting of multiple markets in Finland.
 \\
 The rest of the paper is organised as follows. Section 2 reports current literature in frequency reserve markets, highlighting the need for a bidding optimisation framework for multi-market scenarios. Section 3 presents the study of frequency reserves markets in several countries, leading to the generalised frequency reserves market model. The three bidding strategies are delineated in Section 4, and Section 5 presents the AI-based bidding optimisation framework that implements these strategies using  uncertainty metrics based on supervised and unsupervised machine learning techniques. Section 6 explicates the empirical evaluation on a real-world frequency reserves setting of multiple ancillary markets. Section 7 concludes the paper.

\section{Related work}

This section begins with a deliberation of recent research on frequency reserves of renewable energy sources, highlighting a research gap in bidding strategies for multiple frequency reserve markets. Subsequently, a critical analysis of the closest state of the art in frequency reserve markets is presented. The final subsection reports a review of time-series forecasting methods and uncertainty metrics that can be utilised to develop an AI-based bidding optimisation framework that addresses this gap.
\subsection{Review of related work in frequency reserves}
Traditionally, frequency reserves have been provided using fossil fuel based spinning reserves. Such reserves could be traded on either primary, secondary or tertiary markets as long as the reserve meets the market requirements on minimum capacity and response time \cite{Oyj}. 

However, recent years have seen a surge of innovations adapting renewable energy sources, battery storages and smart loads for the provision of frequency reserves. 
 
Despite this surge of innovations, there has been limited research conducted on the diverse characteristics of the frequency reserve markets and the requirement for participation in these markets. Electric vehicles are used as an example in this paper, as they are targeted by many of the research efforts in frequency reserves, e.g. \cite{yilmaz2012review,suul2016virtual}. While some authors do not explicitly identify the type of reserve being targeted by their solution \cite{debbarma2018grid}, most works are focused either on primary or secondary frequency reserves. The works referenced in this section do not specifically reference the technical requirements of any specific reserves market with the exception of Bokhari et al. \cite{bokhari2017techno}, who consider technical requirements such as response time for several European and North American markets. 
Several authors have proposed electric vehicle charging solutions for primary frequency reserves. In \cite{mizuta2018synthesis}, it was specifically recognize that a sufficiently fast response time is key to participation in primary rather than secondary reserves, but their solution treats the electric vehicle battery as if it were dedicated for the provision of reserves rather than driving. Studies that discuss the design of the frequency controller to meet the response time requirements for primary frequency reserves \cite{izadkhast2017design}, considers the problem of achieving the primary frequency reserve response requirements within the framework of the incentives paid to vehicle owners \cite{moghadam2015distributed}, and recognizes that a single electric vehicle does not have sufficient capability or availability to meet the minimum capacity requirements, and that aggregation is needed for this purpose \cite{rey2018strengthening} have been reported in recent literature. 
A step further, the technical requirements of primary frequency reserves to achieve smooth frequency control \cite{zecchino2019enhanced}, and an aggregation solution aiming to minimize the utilization of the electric vehicle batteries \cite{kariminejad2017hierarchical}, have also been proposed.
\\
A further body of research has been reported on electric vehicle charging solutions for secondary reserves. Falahati et al. \cite{falahati2017grid} have studied the role of electric vehicles within a power system that has wind power generation, but does not consider how the battery capacity is divided between driving and the provision of frequency reserves. Another study \cite{ota2011autonomous} assumes that vehicle usage requirements are known in advance, so that they can be treated as constraints to the frequency control problem. A multi-objective optimization solution has also been presented \cite{kaur2017multiobjective} to balance the trade-off between providing the frequency reserves and satisfying the EVs energy demands under battery degradation constraints. None of these authors discuss response time requirements or the number of vehicles required to meet the technical requirements of the reserves. The approach proposed in \cite{ota2011autonomous} is claimed to be applicable to both primary and secondary reserves, but they do not discuss how it should be adapted to the different technical requirements in these reserves. The other authors do not assess the applicability of their solution for primary or tertiary reserves. In summary, the lack of research on tertiary reserves is evident, and none of the reported approaches take into account the changing price of the secondary reserves. 

\subsection{Analysis of the closest state of the art}
Ferruzzi et al. \cite{ferruzzi2017bidding} consider the problem of bidding on two markets: a spot market and a frequency reserve market. The paper identifies three types of frequency reserves (primary, secondary and tertiary) and their US counterparts (regulation, spinning and not spinning reserve). Only one reserve market is considered in the proposed bidding optimization solution, namely spinning reserve. Further, the proposed bidding optimization solution assumes that the reserve market price is known at the time of placing the bids, although this assumption is not valid for real world reserve markets, in which the market price is determined by an auction mechanism that is executed after bidding is closed \cite{giovanelli2018exploiting}.\\
Rey et al. \cite{rey2018strengthening} consider the problem of bidding aggregated reserve resources on a frequency reserve market. Specifically, the types of resources considered are electricity loads in buildings, such as electric vehicle chargers and heating, ventilation and air conditioning equipment. All of these resources have a primary function other than providing frequency reserves, which constrains the use of these resources on frequency reserve markets. These constraints are specifically related to the maximum power consumption adjustment that the reserve resource can undertake in response to a grid frequency deviation, without violating requirements related to the primary function of the resource. It is assumed that the reserve market price for the upcoming bidding period is known or that it can be easily and accurately predicted from previous market clearings; this assumption is not valid according to an analysis of frequency reserve market price behavior \cite{giovanelli2018exploiting}. The option to trade on more than one frequency reserve market is not considered. It is also assumed that the adjustable power capacity will be the same for all of the hourly bids of the upcoming day-ahead bidding period, due to the policy in Switzerland and some other countries, which are not identified in the paper. However, this policy is not generally in place in Europe, and especially not in Finland, which is the case study for this paper. Thus, our paper considers the possibility to reschedule the utilization of reserve resources to achieve higher adjustable power capacity for the hours in which peak prices are predicted. \\
Perninge \& Eriksson \cite{perninge2017optimal} consider specifically tertiary frequency reserve markets in Northern Europe; the Finnish mFRR market in the case study of this paper is such a market. The objective of the paper is to propose a scheme for the tertiary reserve market operator (i.e. the transmission system operator) to optimally operate the reserve resources that the owners had bid on the tertiary reserve market for a specific hour. This is in contrast to the aims of our paper in two ways. Firstly, the possibility to use the resource on other reserve markets is not considered. Secondly, the profitability for the resource owner is not considered. \\
In summary, all of these state of-the-art works (\cite{ferruzzi2017bidding,rey2018strengthening,perninge2017optimal}) are based on rigorous mathematical models, with simplifying assumptions. Due to these simplifications, these research studies neglect the possibility of capitalizing on high price peaks on frequency reserve markets, either by trading on several frequency reserve markets or by rescheduling adjustable power capacity to peak price hours. Thereby, we propose an alternative bidding optimization approach based on data-driven predictions and AI-based uncertainty metrics to address such assumptions and simplification. \\
Beyond frequency reserve markets, in the broader context of energy markets, some research has been done on approaches that consider multi-stage markets such as day-ahead and hour-ahead, but the aims, assumptions or approach are different from the research presented in this article. Lahouar \& Ben Hadj Slama \cite{lahouar2016comparative} forecast electricity price on half-hour-ahead and day-ahead markets, but do not act on the forecasts. Caramanis et al. \cite{caramanis2010load} propose a bidding strategy on hour-ahead and day-ahead electricity markets, but do not address the problem of forecasting the prices or using the forecasts to improve the bidding. Reddy \& Singh \cite{reddy2018optimal} propose a demand response strategy on day-ahead and hour-ahead markets, but assume that the prices of the upcoming bidding period are known. Shafie-khah et al. \cite{shafie2014optimal} propose scheduling of wind power production on multi-stage markets; instead of trying to predict market price, the price uncertainties are handled by the Monte Carlo method. Zhong \cite{zhong2015multi} propose a demand response scheme for trading in two settlement markets; the benefit for the market operator is reduced market clearing prices, whereas our work aims at increasing revenues for market participants. Zhao et al. \cite{zhao2019look} propose a trading scheme for a virtual power plant operating on multi-stage electricity markets, so that market price uncertainly is handled with scenario trees. Böhringer et al. \cite{bohringer2019trading} trade a factory’s flexible energy consumption capability on multi-stage electricity markets, in which market price uncertainty is treated as a risk without trying to forecast the price. 
None of these approaches that focus on multi-stage markets have addressed the challenges of price prediction with uncertainty in volatile markets for generating a profit from price peaks. The  focus of this paper is to address these challenges using an AI based bidding optimization framework that implements three bidding strategies that encompass real-time market conditions and data-driven predictions of price.

\subsection{Time-series forecasting and uncertainty metrics}
Price forecasting in energy markets has  become a fast growing interest with the introduction of competitive markets.\cite{karthikeyan2013review} There are five main categories of time-series forecasting methods use for price prediction:  multi-agent models or game theoretic models\cite{giovanelli2018exploiting}, fundamental methods, reduced form models, statistical models and computational intelligence models \cite{weron2014electricity}, with statistical and computational intelligence methods are the most effective in practice. \cite{yang2019hybrid}\par
\noindent Variations of statistical models like autoregressive moving average
(ARMA) \cite{liu2013applying}, autoregressive \cite{conejo2005day} integrated moving average (ARIMA) are commonly used for electricity price forecasting.
In rapidly-changing markets, statistical models are ineffective and computational intelligence methods outperform others\cite{chan2012load}. These computational intelligence methods include Support Vector Regression Machines (SVR)\cite{drucker1997support} and Artificial Neural Network (ANN) methods. 
ANN based methods have been accurate in predicting ancillary market prices with high variability\cite{giovanelli2018exploiting}. Despite this accuracy of prediction, the high variability has led to uncertainty mainly due to the data scarcity and volatility of energy markets. 
Addressing this uncertainty of prediction has been researched in several other domains, and it can be theorised into two aspects, aleatory uncertainty and epistemic uncertainty \cite{hester2012epistemic}. Aleatory uncertainty is an uncertainty due to chance, also known as stochastic uncertainty, and it cannot be resolved with more data. In contrast, epistemic uncertainty can be considered as a source of uncertainty due to lack of knowledge of the environment \cite{fox2011distinguishing}, and it can be addressed with more data. e.g: \cite{mobiny2019risk}
\\
Uncertainty metrics are necessary when the data is scarce or there is inherent randomness in the process generating the data.
Bayesian inference based methods have been traditionally used to measure uncertainty \cite{mobiny2019risk}. Bayesian neural networks (BNN) is a variant of neural networks used for the purpose of uncertainty estimation. A BNN generates a distribution over the output for a given input, in contrast to Artificial neural networks (ANN), which produce a point output for given input\cite{neal1993bayesian}. However, this variational inference is a costly process\cite{mobiny2019dropconnect}. Recent advances in BNN approximation with dropouts give efficient estimates of uncertainty in neural networks. This is commonly known as Monte Carlo dropout (MC-dropout).\cite{gal2016dropout}.

\section{Generalised model of frequency reserves market}

Frequency reserves are divided into primary, secondary and tertiary reserves. Primary reserves are often called Frequency Containment Reserves (FCR), which are responsible for stabilizing the frequency rapidly after a frequency deviation occurs. Secondary and tertiary reserves are often called Frequency Restoration Reserves (FRR). FRR is activated automatically for secondary reserves (aFRR) and manually for tertiary reserves (mFRR). The terms FCR, aFRR and mFRR are broadly used in Europe, but they are not globally established. \cite{bokhari2017techno} present a thorough comparison of North American and European markets. A major research challenge related to predicting frequency reserves markets is related to generality: there is considerable diversity between the terminology, market rules and technical specifications between countries. The following factors should be taken into account in order to improve the generality of solutions (examples are based on analyzing the ENTSOE-E Transparency Platform https://transparency.entsoe.eu/ ):
\begin{itemize}
    \item Bidding interval: how often are auctions held? As an example of a week-ahead market, German FCR has the same price for the entire week and the auctions are held once a week to determine the price of the next week. As an example of a day-ahead market, in Finnish FCR the auctions are held daily and 24 different prices are determined for the hours of the next day. As an example of an hour-ahead market, in Finnish mFRR the auctions are held hourly to determine the price for the upcoming hour.
    \item Market changes: what was said in the previous bullet applied to the German FCR market until 2018. In 2019 the price changes daily.
    \item Number of markets: many European countries have one single FCR market, but Northern European countries have separate FCR markets for normal operation (FCR-N) and disturbances (FCR-D).
    \item Epochs: the most typical epoch is one hour, but e.g. French FCR, aFRR and mFRR have a 30min interval.
    \item Symmetry: in a symmetric market, participants must provide both down and up regulation. In an asymmetric market, these are separate markets with separate prices, and it is not required to participate on both. FCR are often symmetric and aFRR are often asymmetric, e.g. as is the case of Finland, but as one exception French aFRR is symmetric.
    \item Peak hours: some markets distinguish between peak and off-peak hours, but the details are country specific. For example in Finnish aFRR, the market is held only on peak hours on weekdays, with auctions determining a separate price for each hour. In German aFRR, the price changes every 4 hours.
\end{itemize}
\begin{table*}[ht]
    \centering
    \resizebox{\textwidth}{!}{
    \begin{tabular}{|p{3cm}|p{4cm}|p{5cm}|p{3cm}|}\hline
        \parbox{1.5cm}{\textbf{Reserve type/synonyms}} & \textbf{Finland \cite{Oyj2019b}} & \textbf{Germany} & \textbf{France} \\ \hline
        Primary/ Frequency containment/ regulation &There are two primary reserves markets FCR-N and FCR-D, operated within the geographical area of Finland by the TSO Fingrid. FCR-N is symmetric and FCR-D is activated only in cases of under-frequency&\multicolumn{2}{|p{8cm}|}{There is a single primary reserve market FCR in the Continental Europe Synchronous Area for primary frequency control (which excludes Finland, Sweden, Norway). The market is symmetric \cite{frequencyancillaryconditions}. The mechanism for contracting reserves is cross-border tenders at www.regelleistung.net}	\\ \hline
        Secondary/ automatic frequency restoration/ spinning &	Asymmetric aFRR market operated by Fingrid &\multirow{2}{*}{\parbox{5cm}{Asymmetric aFRR and mFRR internal markets jointly operated by Austrian TSO APG and the four German TSOs 50Hertz, Amprion, TenneT and TransnetBW \cite{amrioncontrolenergy} }} &Symmetric aFRR market operated by RTE \cite{frequencyancillaryconditions} \\ \cline{1-2} \cline{4-4}
        Tertiary/ manual frequency restoration/ not spinning &	Asymmetric mFRR market operated by Fingrid & & \parbox{3cm}{Asymmetric mFRR market operated by RTE \cite{frequencyancillaryconditions}.}\\ \hline 
    \end{tabular}
    }
    \caption{A comparison of how frequency reserves markets are organized in three European countries}
\end{table*}

Currently, an EU wide harmonization of frequency reserves markets is underway. Key observations from this harmonization include shortening of bidding intervals and epochs and moving from weekly to day-ahead markets \cite{enotso_1}.
Table 2 defines the terms used throughout this paper. Symbols and abbreviations are defined only when necessary to aid the formalization of the proposed methodology.
Figure 1 shows a UML (Unified Modeling Language) class diagram that defines the relationships between the key concepts in Table 2. The class diagram is limited to capturing the relationships between the concepts used in this paper. It does not have the detail needed to support a virtual power plant implementation. Figure 2 presents a generic operating model of a frequency reserves market, using terms and symbols from Table 2 to parameterize the characteristics that vary across the primary, secondary and tertiary reserves of the analyzed countries.

\begin{table*}[ht]
    \centering
    \resizebox{\textwidth}{!}{
    \small{
    \begin{tabular}{p{3cm}|p{3cm}|p{10cm}}\hline
        Symbol/Abbreviation& Term& Definition \\ \hline
        TSO & Transmission System Operator& \\ \hline
        &Reserve market &		One of several markets in which a Transmission System Operator procures controllable power production or consumption capacity that can be activated in the event of grid disturbances resulting in the grid frequency deviating from the nominal  \\ \hline
        M &	Reserve markets &	Set of reserve markets considered by the bidding strategy  \\ \hline
        $M_{day-ahead}$ & Day-ahead reserve markets&	Subset of M in which the bidding interval is one day. $Bid_{DL}^m$ for $m \in M_{day-ahead}$ is on the previous day   \\ \hline
        $m_{epoch-ahead}$ & Epoch-ahead reserve markets&	$m_{epoch-ahead}$ of M in which the bidding interval is one epoch. $Bid_{DL}^m$ is on the previous epoch. It is assumed there is only one such market in M.   \\ \hline
	    & Bidding interval &	The interval between two successive auctions that are held for a specific reserves market. E.g. for day-ahead markets the interval is one day and for hourly markets, the interval is one hour   \\ \hline
        $A_i$ &	Auction&	An auction that is held for bidding interval “i” of a specific reserve market in which a TSO procures the necessary amount of reserves by accepting or rejecting bids submitted to that auction   \\ \hline
        $Bid_{DL}^m$ &	Bidding deadline &	The deadline by which bids must be submitted in order for them to be considered for a specific auction on market “m”.  \\ \hline
	    & Reserve Unit &	A unit that fulfils the requirements concerning the providing of the reserve. A Reserve Unit can consist of one more Reserve Resources.  \\ \hline
	    & Reserve Resource &	an individual resource capable of control; a power plant, consumption facility or energy storage facility.  \\ \hline
	    & Epoch &	The shortest time period for which a reserve unit can be offered on a frequency reserves market  \\ \hline
        $N_{epochs}$ &	Epochs per interval &	The number of epochs in one bidding interval   \\ \hline
        $S_P$ & 	Service Provider &	a party that has a valid agreement on the delivery of reserve services.  \\ \hline
        $C[e]$ &	Capacity &	Capacity is the power consumption/production that can be controlled in response to a frequency deviation according to the technical specifications of the reserve market. $C[e]$ is the total untraded capacity of all reserve resources owned by the Service Provider.   \\ \hline
        $C_{fee}^m[e]$ &	Capacity  fee &	refers  to  the  compensation  paid  by  the TSO to the Service Provider for the maintaining of the reserve. The fee is specific to epoch “e” on market “m”   \\ \hline
        $Proc_i^m[]$ &	Procured capacity &	 $Proc_i[e]$ is the capacity procured by the TSO for each epoch “e” in the bidding interval “i” on market “m”   \\ \hline
        $R_fee$ &	Requested fee &	The minimum Capacity fee at which the Service Provider is willing to provide a Reserve unit  \\ \hline
	    &Bid &	For a specific reserve market, the bid specifies the capacity that can be provided for a specific epoch and the Requested fee at which the service provider is willing to provide the capacity  \\ \hline
	    & Order &	A bid that was accepted by the TSO  \\ \hline
        $F_{fee}^m[e]$ &	Forecasted capacity fee &	Forecasted capacity fee for epoch “e” on market “m”  \\ \hline
        $UF_{fee}^m[e]$ &	Uncertainty of capacity forecast &	Uncertainty for $F_{fee}[e]$ on market “m”   \\ \hline
        $UT_m$ &	Uncertainty threshold &	Uncertainty threshold for reserve market “m”. If $UF_{fee}^m[e] < UT_m, F_{fee}^m[e]$ is considered reliable.   \\ \hline
        $MPP$ &	Minimum profitable price &	The minimum Capacity fee at which the Service Provider can profitably provide a Reserve unit to a reserve market. This differs from $R_{fee}$ in the sense that the Service Provider may set $MPP >= R_{fee}$ depending on its bargaining strategy  \\ \hline
        $T_{comp}$ &	Computing time &	Time required to compute $UF_{fee}^m[e], \;UT_m, \; F_{fee}^m[e]$  \\ \hline

    \end{tabular}
    }
    }
    \caption{Terms and Abbreviations used in this article}
\end{table*}

\begin{figure}[H]
	\centering
		\includegraphics[scale=.09]{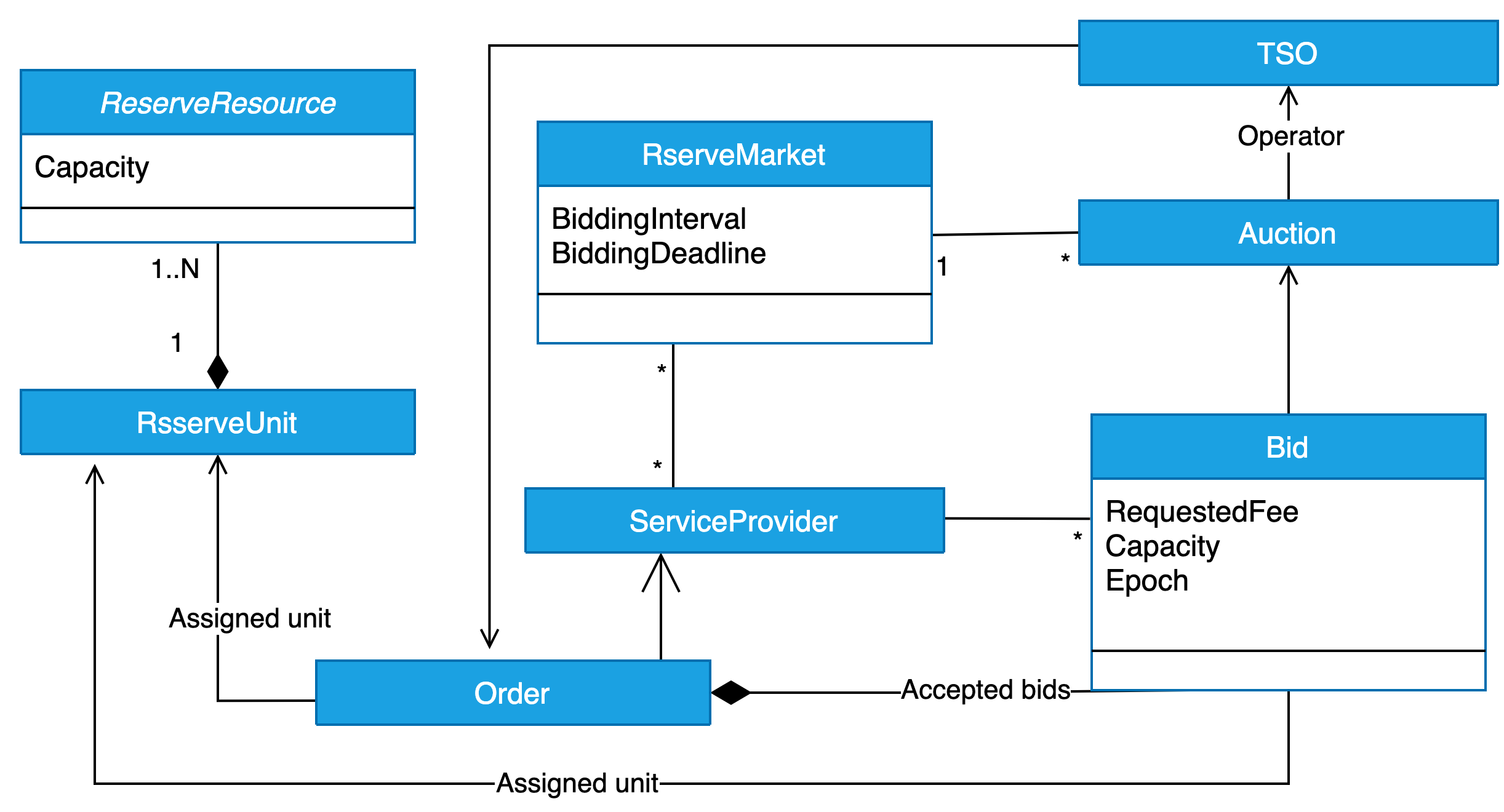}
\caption{UML class diagram of generalised frequency reserves market \cite{Oyj2019b}. 
\\ Refer Table 2 for terms and abbreviations}
\end{figure}

\begin{figure}[H]
	\centering
		\includegraphics[scale=.20]{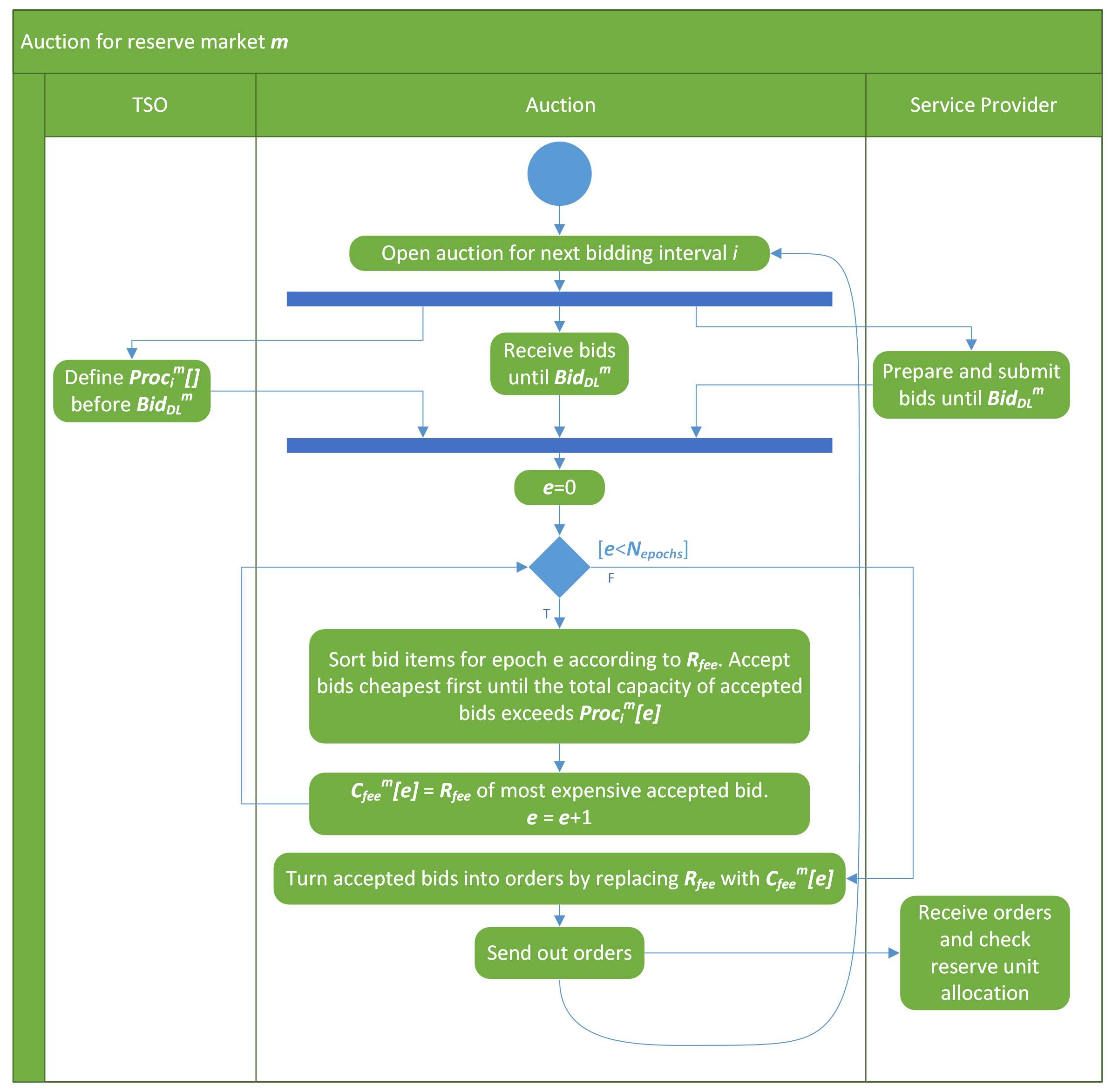}
\caption{UML activity diagram of generalised frequency reserves market.
\\Refer Table 2 for terms and abbreviations}

\end{figure}

\section{Bidding strategies}

The proposed bidding strategies are optimised using AI based market forecasts as well as uncertainty metrics of these forecasts. The rationale for this integration of technologies is as follows. As discussed in section 2, AI has emerged as a state-of-the-art approach for time series forecasting in the energy domain; however, none of the state-of-the-art research studies have forecasted price peaks with high confidence, and this is the motivation for developing the uncertainty metric. In contrast to mathematical optimization, bidding strategies based on price prediction and uncertainty metrics can incorporate knowledge of the trading opportunities on the multi-stage markets and the possibilities for rescheduling frequency reserve resources.
Based on the analysis of global frequency reserves markets and the generalised model developed in the previous section, the following assumptions are broadly valid:
\begin{enumerate}
    \item The primary and secondary reserves markets are day-ahead markets. In several countries there are two different primary reserves markets, so there are several day-ahead markets to choose from.
    \item There is one tertiary reserves market, which is an epoch ahead market.
\end{enumerate}
In the following subsections, three bidding strategies are proposed. Strategies 1 and 2 address the scenario of non-reschedulable reserve resources whereas strategy 3 addresses the scenario of reschedulable reserve resources. The difference between strategies 1 and 2 is in the way that they make decisions whether to trade on day-ahead or hour-ahead frequency reserves markets. In this paper, the profitability of strategy 1 versus strategy 2 is evaluated empirically with a case study. Strategy 1 tries to trade all of the available reserve resource capacity on day-ahead markets, and participates on hour-ahead markets only when bids for day-ahead markets were rejected. Strategy 2 refrains from bidding on day-ahead markets if the hour-ahead markets are forecasted to have a higher price. Strategy 3 reschedules the utilization of reschedulable reserve resources to hours on which high reserve market prices have been forecasted.

\subsection{Bidding strategy 1}
The acceptance decision for day ahead bids will come on the same day as when the bids were placed e.g. \cite{caramanis2010load,lahouar2016comparative}, so it is reasonable to formulate a two-phase bidding strategy. First bid will be on day-ahead markets, after these bids have been accepted or rejected, bid the remaining capacity on epoch-ahead markets. Algorithm 1 presents a straightforward strategy that selects, for each epoch, the day-ahead market with the highest price forecast. Only markets with reliable forecasts are considered. After the acceptance or rejection information for these bids have been received, the capacity of the rejected bids is offered onto the epoch ahead market, i.e. the tertiary reserves market. From Figure 2 it can be seen that all accepted bids receive the same capacity fee regardless of what the requested fee was. In this kind of auction, it is reasonable to make bids at the lowest requested fee which is better than no-deal. The concept of MPP was defined for this purpose. The actual value of MPP depends on opportunity costs resulting from other possible usage scenarios of the reserve resource. It also depends on any usage cost of the resource as it participates in the reserve: for example, if the resource is a battery, participating in frequency reserves will involve charging and discharging operations that would shorten the lifespan of the battery. Quantification of the MPP is thus specific to the type of reserve resource and the system that it belongs to, so such calculations are beyond the scope of this article.

\begin{algorithm}[t]
\DontPrintSemicolon
\KwIn{ All $m\in M_{day-ahead} \;\&\; M_{epoch-ahead}$ }
\KwOut{ For each $e$  market $m$ to bid on $MPP$}
Wait until $time \gets earliest\: Bid_{DL}^m \:of \: m\in M_{day-ahead}$ - $T_{comp}$\;
\For{each $m \in M_{day-ahead}$}
{
    Compute $F_{fee}^m[e], UF_{fee}^m[e]$ and $UT^m$
}
$e \gets 0$\;
\While{$i \leq  N_{epochs} $}{
  $M' \gets \emptyset$\;
  \For{each $m \in M_{day-ahead}$}
  {
  \uIf{$UF_{fee}^m[e] \leq UT^m$}
  {
    $M' \gets M'\cup m$\;
  }
    Select $m \in M'$ with highest $F_{fee}^m[e]$\;
    Bid on $m$ for epoch $e$ with price $MPP$ and capacity $C[e]$\;
  }
  $e \gets e + 1$\;
}
Wait for market clearance and update $C[e]$ to match total capacity of rejected bids\;
\For{$e$ with $C[e] >0$}
{
    Bid with $MPP$ on $m_{epoch-ahead}$\;
}
\caption{{\sc Bidding Strategy 1, check Figure A.10 for more information}}
\label{algo:Bidding Strategy_1}
\end{algorithm}
\subsection{Bidding strategy 2}
Bidding strategy 1 will not perform well if the tertiary reserves market has a higher price than any other market for a specific epoch. In such a scenario, it would be more profitable to refrain from bidding on any day-ahead market and to bid all the capacity on the tertiary market. Bidding strategy 2 in Algorithm 2 implements this reasoning.

\begin{algorithm}[t]
\DontPrintSemicolon
\KwIn{ All $m\in M_{day-ahead} \; \& \; M_{epoch-ahead}$ }
\KwOut{ For each $e$  market $m$ to bid on $MPP$}
Wait until $time \gets earliest\: Bid_{DL}^m \:of \: m\in M_{day-ahead}$ - $T_{comp}$\;
\For{each $m \in M_{}$}
{
    Compute $F_{fee}^m[e], UF_{fee}^m[e]$ and $UT^m$
}
$e \gets 0$\;
\While{$i \leq  N_{epochs} $}{
  $M' \gets \emptyset$\;
  \For{each $m \in M_{day-ahead}$}
  {
  \uIf{$UF_{fee}^m[e] \leq UT^m$}
    {
    $M' \gets M'\cup m$\;
   }
      Select $m \in M'$ with highest $F_{fee}^m[e]$\;
      \If{$F_{fee}^{m\_epoch-ahead}[e] > F_{fee}^m[e]$}
      {
        continue\;
      }
      Bid on $m$ for epoch $e$ with price $MPP$ and capacity $C[e]$\;
  }
  $e \gets e + 1$\;
}
Wait for market clearance and update $C[e]$ to match total capacity of rejected bids and capacity that was not bid previously\;
\For{$e$ with $C[e] >0$}
{
    Bid with $MPP$ on $m_{epoch-ahead}$\;
}
\caption{{\sc Bidding Strategy 2, check Figure A.11 for more information}}
\label{algo:Bidding Strategy_2}
\end{algorithm}

\subsection{Bidding strategy 3}
Previously presented bidding strategies assumed non-reschedulable reserve resources, i.e. a fixed $C[e]$. Bidding strategy 3 focuses on reschedulable reserve resources, where the task is to perform the rescheduling in order to determine the $C[e]$ that will be traded on reserve markets for each epoch.
\\
As an example, consider an electric vehicle (EV) that is plugged in all night to a home charger, which is able to fully charge the EV at 22kW in 2h. The EV needs to be fully charged by 8AM, so a charging of 44kWh needs to be scheduled. If the charger has a frequency control feature, it can be used on frequency reserves markets during the hours that it charges the EV: it is possible to simply curtail the charging current in the event of frequency deviations. Starting from midnight, it is possible to participate in day-ahead reserve markets. Several charging schemes are possible:
\begin{enumerate}
    \item Charge at a constant power for the 8 hour period from midnight to 8AM, resulting in a charging power of 44kWh/8h = 5,5W.
    \item Charge the EV in two hours starting at midnight.
    \item Charge the EV in two randomly chosen one-hour periods starting earliest at midnight and ending latest at 8AM.
    \item Pick the two one hour periods with highest predicted price and charge the EV at 22kW during those hours. The periods do not need to be consecutive.
\end{enumerate}
The charging power on epoch $e$ is the power that can be traded at that epoch, i.e. $C[e]$.
\subsubsection{General formulation}
 Table 3 defines additional terms that are only relevant for reschedulable reserve resources.
\\
\begin{table*}[ht]
    \centering
    \resizebox{\textwidth}{!}{
    \begin{tabular}{p{3cm}|p{3cm}|p{10cm}}\hline
        Symbol/Abbreviation& Term& Definition \\ \hline
        $e_h$ &	Epochs per h &	The number of epochs per hour \\ \hline 
        $P_{max}$ &	Maximum power of reschedulable resource &	The maximum power (kW) at which the reschedulable reserve resource can be operated  \\ \hline
        $e_{earliest}$ &	Earliest epoch &	Earliest epoch at which the resource can be scheduled  \\ \hline
        $e_{latest}$ &	Latest epoch &	Latest epoch at which the resource can be scheduled \\ \hline
        $e_{total}$ &	Total epochs &	Number of epochs in the range $e_{earliest} .. e_{latest}$  \\ \hline
        $E$ &	Total reschedulable energy &	Required total reschedulable energy consumption or production between $e_{earliest}$ and $e_{latest}$ (in kWh) \\ \hline
        $N_{min}$ &	Minimum number of epochs &	The minimum number of epochs required to consume/produce E at $P_{max}$  \\ \hline
        $M_{best}[e]$ &	Best market based on forecast &	The market that had the highest price with acceptable prediction uncertainty on epoch e  \\ \hline
    \end{tabular}
    }
    \caption{Terms and Abbreviation specific to bidding strategy 3}
\end{table*}
A universal model for a reschedulable reserve resource would be a research question in its own right, as it would need to capture aspects such as user preferences, unexpected events such as an EV being plugged out in the middle of the night, as well as device specific constraints such as battery recovery times. In this paper, the example from section 3.4.1 is generalized as follows. We consider a reschedulable reserve resource that will produce or consume a quantity of energy E. This action may be scheduled so that it starts earliest on epoch $e_{earliest}$ and is completed latest on epoch $e_{latest}$, so that the power production or consumption is constant during the epoch. This power is the $C[e]$ that the reserve resource can provide. In this case:
\begin{equation}
N_{min}=roundUp(E/p_{max}) * e_h
\end{equation}
Here $roundUp()$ is a function to convert floating point value to nearest largest integer value.
The schemes 1-4 can be formalized as follows:
\begin{enumerate}
    \item For all e in the range $e_{earliest} .. e_{latest}$, $C[e] = E/( etotal * eh)$.
    \item For all $N_{min}$ consecutive epochs starting at $e_{earliest}$, $C[e] = P_{max}$. For all other epochs, $C[e] = 0$.
    \item For $N_{min}$ randomly chosen epochs in the range $e_{earliest}$ $.. e_{latest}$, $C[e] = P_{max}$. For all other epochs, $C[e] = 0$.
    \item Algorithm 3 shows a bidding strategy that concentrates the reschedulable capacity to the epochs and markets that are expected to have the highest prices. For all $e$ in the range $e_{earliest} .. e_{latest}$, forecast all markets and select the market that had the highest price with acceptable prediction uncertainty, i.e. $M_{best}[e]$. Select $N_{min}$ epochs in the range $e_{earliest} .. e_{latest}$ for which $M_{best}[e]$ had the highest prices and bid $C[e] = P_{max}$ for those epochs. For all other epochs, $C[e] = 0$.
\end{enumerate}

\begin{algorithm}[t]
\DontPrintSemicolon
\KwIn{ All $m\in M_{day-ahead} \; \& \; M_{epoch-ahead}$ }
\KwOut{ For each $e$  market $m$ to bid on $MPP$}
Wait until $time \gets earliest\: Bid_{DL}^m \:of \: m\in M_{day-ahead}$ - $T_{comp}$\;
\For{each $m \in M_{}$}
{
    Compute $F_{fee}^m[e], UF_{fee}^m[e]$ and $UT^m$
}
$e \gets e_{earliest}$\;
\While{$i \leq  e_{latest} $}{
  $M' \gets \emptyset$\;
  \For{each $m \in M_{day-ahead}$}
  {
  \uIf{$UF_{fee}^m[e] \leq UT^m$}
    {
    $M' \gets M'\cup m$\;
   }
      Select $m \in M'$ with highest $F_{fee}^m[e]$\;
      $M_{best}[e] \gets m$\;
      \While{$i < N_{min}$}
      {
        $i \gets 0$\;
        \If{ $e \in \; range(e_{earliest},e_{latest}) $}
        {
            \If{$e$ without a bid \& $F_{fee}^{M\_best[e]}[e] \; is \; highest$}
            {
                Bid on $M_{best}[e]$ for epoch e with price $MPP$ and capacity $P_{max}$
            }
        }
        $i \gets i+1$\;
      
      }
  }
  $e \gets e + 1$\;
}
Wait for market clearance and update $C[e]$ to match total capacity of rejected bids and capacity that was not bid previously\;
\For{each rejected bid}
{
    make the same bid to $m_{epoch-ahead}$\;
}
\caption{{\sc Bidding strategy with reschedulable loads, check Figure A.12 for more information}}
\label{algo:Bidding_strategy_with_reschedulable_loads}
\end{algorithm}

Schemes 1-3 are straightforward and will be implemented in order to provide a point of comparison to the performance of scheme 4.
\section{The AI-based bidding optimization framework}

The AI-based bidding optimization framework implements the three bidding strategies delineated above, the first two strategies for non-reschedulable loads, and the third for reschedulable loads with four alternative rescheduling schemes. The following subsections focus on the constituents of this framework, price prediction, supervised and unsupervised uncertainty metric calculation, and uncertainty metric evaluation. 

\subsection{Price Prediction}
Price prediction is based on a three-layer neural network approach proposed  in \cite{giovanelli2018exploiting}, where $n=64$ with hidden layer of 32 and output of 24 corresponding to 24 hours. An optimum window size of 180 days was used. The three-layer neural network was converted into a Bayesian neural network for  point prediction of 24 hour output as averaged over $N=500$ iterations with the same dropout rate 40\% at the testing time.

\subsection{Uncertainty metric}
Despite using AI, the prediction of price can still be uncertain due to the scarcity of data and volatility of renewable energy sources\cite{gonzalez2015impact,liu2019worst}. It is important to quantify the uncertainty of the price prediction for optimised bidding. We used a combination of two uncertainty measures, supervised and unsupervised for better representation of price prediction uncertainty in frequency reserve markets. For the former, we used MC-Dropout, a well-established uncertainty measure for deep learning \cite{gal2016dropout}. For the latter, we introduced a novel unsupervised uncertainty measure which will independently evaluate the training data for epistemic uncertainty.

\subsubsection{Supervised uncertainty metric}
As the supervised uncertainty metric, we used the MC-Dropout method with the neural network evaluated $N$ times for a single price prediction. This resulted in a distribution of predictions $\hat{p}^1_t,\hat{p}^2_t...\hat{p}^N_t$ for a single price prediction $\hat{p}_t$. The prediction $\hat{p}_t$ and the normalized uncertainty $nu_t$ are evaluated from the distribution as follows:
\begin{equation}
\hat{p}_t = \frac{1}{N}\sum_{j=1}^{N} \hat{p}^j_t
\end{equation}
\begin{equation}
u_t = \sqrt{\frac{1}{N-1}\sum_{j=1}^N(p^j_t-\hat{p}_t)^2}
\end{equation}
\begin{equation}
nu_t = \frac{u_t}{\hat{p}_t}
\end{equation}
\subsubsection{Unsupervised uncertainty metric}
We introduce a novel unsupervised uncertainty metric that is independent of the three-layer neural network used for prediction. 
The unsupervised uncertainty metric is based on the Growing Self Organizing Map (GSOM) algorithm. The GSOM algorithm  is an unsupervised neural network variant of Self organizing maps(SOM), which starts with four nodes and grows nodes to represent the input data\cite{alahakoon2000dynamic}. It has been used in many unsupervised applications related to clustering, dimensional reduction and abstract representation in the energy domain \cite{de2011data, de2016data, jayaratne2019unsupervised}.
The GSOM algorithm generates a topological latent representation of the input vectors that can be utilised to identify clusters of nodes based on a distance measure. We evaluated the GSOM algorithm alongside the SOM algorithm, T-SNE, and K-means algorithm (all of which generate similar topological maps based on unsupervised learning). The GSOM algorithm possesses the most effective topology preserving measures for a non-linear dataset such as the data streams in this scenario. 
This uncertainty metric based on the unsupervised learning capability of the GSOM algorithm generates an independent assessment of data uncertainty. In contrast, MC-Dropout generates an uncertainty value for both the data and the predictive model. If both MC-Dropout uncertainty and GSOM uncertainty values are similar, it is an indication that uncertainty for the prediction is from lack of knowledge in the data.
\\
In terms of an uncertainty metric, when input vectors clustered to a GSOM node have low variance, it is an indication that price is accurately represented by these input vectors and thereby epistemic uncertainty is low for those values. In contrast when a GSOM node has high variance, this is an indication that the price is not completely represented by the data vector and epistemic uncertainty is high. Further, in the absence of any similarity, it is an entirely new pattern with high uncertainty. Using the above notion we define the GSOM uncertainty for price prediction as follows.
If a new data point is clustered to a node with previous price points $P_1,P_2..P_c$ with mean price of $\mu$
\begin{equation}
u_t = \sqrt{\frac{1}{N-1}\sum_{j=1}^c(p_j-\mu)^2}
\end{equation}
If a new data point is clustered to a node with no previous price point
\begin{equation}
u_t = \infty
\end{equation}

To obtain the normalized uncertainty value $u_i$ is normalized using Equation 3.

\subsection{Uncertainty evaluation metric}

MC-dropout and GSOM uncertainty metrics provide estimation for uncertainty, but there are no ground truth values for model prediction, and it is challenging to quantitatively evaluate the uncertainty values\cite{mobiny2019risk}. We extend the evaluation metric proposed in \cite{mukhoti2018evaluating,mobiny2019risk} for categorical value prediction, which will capture the properties we require our model to satisfy. The method is based on the assumption that the model is robust when it is accurate and also uncertain if the model is inaccurate. To calculate this metric we need to convert our predictions to binary classification from continuous values of price prediction and uncertainty value. 
We proposed the following approach for converting a price prediction for binary classification (accurate or inaccurate) in the training dataset:
We consider $S$ markets and for each hour t we predict price $\hat{p}^s_t \; s \in S$ and actual price $p^s_t \; s \in S$
Consider prediction is accurate if:
\begin{equation}
accurate  \gets maxMarket(\hat{p}^s_t) =maxMarket(p^s_t)
\end{equation}
Here $maxMarket()$ is a function which iterate all $s$ in $S$ and find the max priced market.

The certainty of the prediction is determined as follows:
\begin{equation}
certain  \gets nu_t < u_{th}
\end{equation}
Here $u_{th}$ is the uncertainty threshold where $u_{th} \in [0,1]$. $u_{th}$ is a pre-calculated constant value for each market, where optimum $u_{th}$ value for a market is calculated as a maximizing optimization problem of uncertainty accuracy (UA) the exact value was calculated using a hierarchical search algorithm through [0,1] range with a step 0.01
With the above model of classification, we can determine the confusion matrix with following components, accurate and certain ($N_{ac}$, accurate and uncertain ($N_{cu}$), inaccurate, certain ($N_{ic}$) and inaccurate, uncertain ($N_{iu}$). 
We can determine the accuracy and two conditional probabilities $P(accurate|certain)$ and 
\\ $P(inaccurate|uncertain)$ as follows,
\begin{equation}
P(accurate|certain) = \frac{N_{ac}}{(N_{ac}+N_{ic})}
\end{equation}
\begin{equation}
P(inaccurate|uncertain) = \frac{N_{iu}}{(N_{au}+N_{iu})}
\end{equation}
\begin{equation}
Uncertainty \; Accuracy(UA) = \frac{N_{ac}+N_{iu}}{(N_{ac}+N_{au}+N_{ic}+N_{iu})}
\end{equation}

\section{Experiments and results}

\subsection{A real-world frequency reserves setting}

\noindent The proposed AI-based bidding optimization framework was evaluated on the Finnish frequency reserve markets. We considered two day-ahead markets FCR-N and FCR-D and one hour-ahead market mFRR for the case study. We assumed that our resources had a 10MW capacity to react to frquency deviations at all times, so each bid was 10MW for bidding strategy 1 and 2. We used May-June time of the year for the evaluation as it has the most variation in price for FCR-N and FCR-D markets. We selected 30 days starting from 2018-05-10 for the analysis of the bidding optimization with price prediction with uncertainty.\par
\begin{table}[H]
    \centering
    \resizebox{\columnwidth}{!}{
    \begin{tabular}{c|c|c|c}\hline
        &Primary reserve& Secondary reserve& Tertiary reserve \\ \hline
        Finnish markets &FCR-N, FCR-D & aFRR up \& down & mFRR up \& down \\ \hline
        Bidding period & Day-ahead & Day-ahead & Hour-ahead  \\ \hline
        Market interval & 1h & 1h &	1h  \\ \hline
        Symmetric &	Yes & No &	No  \\ \hline
        Peak hours &	No &	Yes &	No  \\ \hline
    \end{tabular}
    }
    \caption{Finnish ancillary services markets}
\end{table}
\noindent\textbf{Primary reserve}: \\
\noindent The two primary frequency reserves markets are: \\
1) FCR-N (Frequency Containment Reserves – Normal Operation) \\
2) FCR-D (Frequency Containment Reserves – Disturbances).
The markets have similar technical requirements, with the main difference being that FCR-D needs to be activated more quickly. The technical requirements \cite{Oyj} are such that many distributed energy resources, such as batteries, could be pre-qualified for both markets. Since bidding is on an hourly basis, and since both markets have the same bidding schedule \cite{Oyj2018}, a Virtual Power Plant(VPP) could choose to bid on either of these markets on an hourly basis, always selecting the market on which the price is expected to be higher.
\\

\noindent\textbf{Secondary reserve:}
\\The aFRR market in Finland is currently only active on certain morning and afternoon hours on weekdays, with very little price fluctuation. The TSO Fingrid intends to change this in the near future\cite{Vanska}. Thus, aFRR was not included in the case study.\\

\noindent\textbf{Tertiary reserve:}
\\Fingrid confusingly uses the terms mFRR and balancing capacity markets interchangeable \cite{Oyj2019b}. This is an hour-ahead market with separate up and down markets like aFRR.

\subsection{Experiment 1:Uncertainty Evaluation}
In this experiment, we predict the price for 30 days  starting from 2018-05-10 with uncertainty evaluation based on both MC-dropout and GSOM methods for the three considered markets: FCR-N, FCR-D and mFRR. Figure 3 visualizes the three markets each sub-figure depict the actual price, predicted price and uncertainty measure. Further, We have used three uncertainty thresholds to identify uncertain price predictions for each market, uncertain price regions are shaded in red for each market.
\begin{figure}[H]
\begin{subfigure}{.5\textwidth}
  \centering
  \includegraphics[scale=.16]{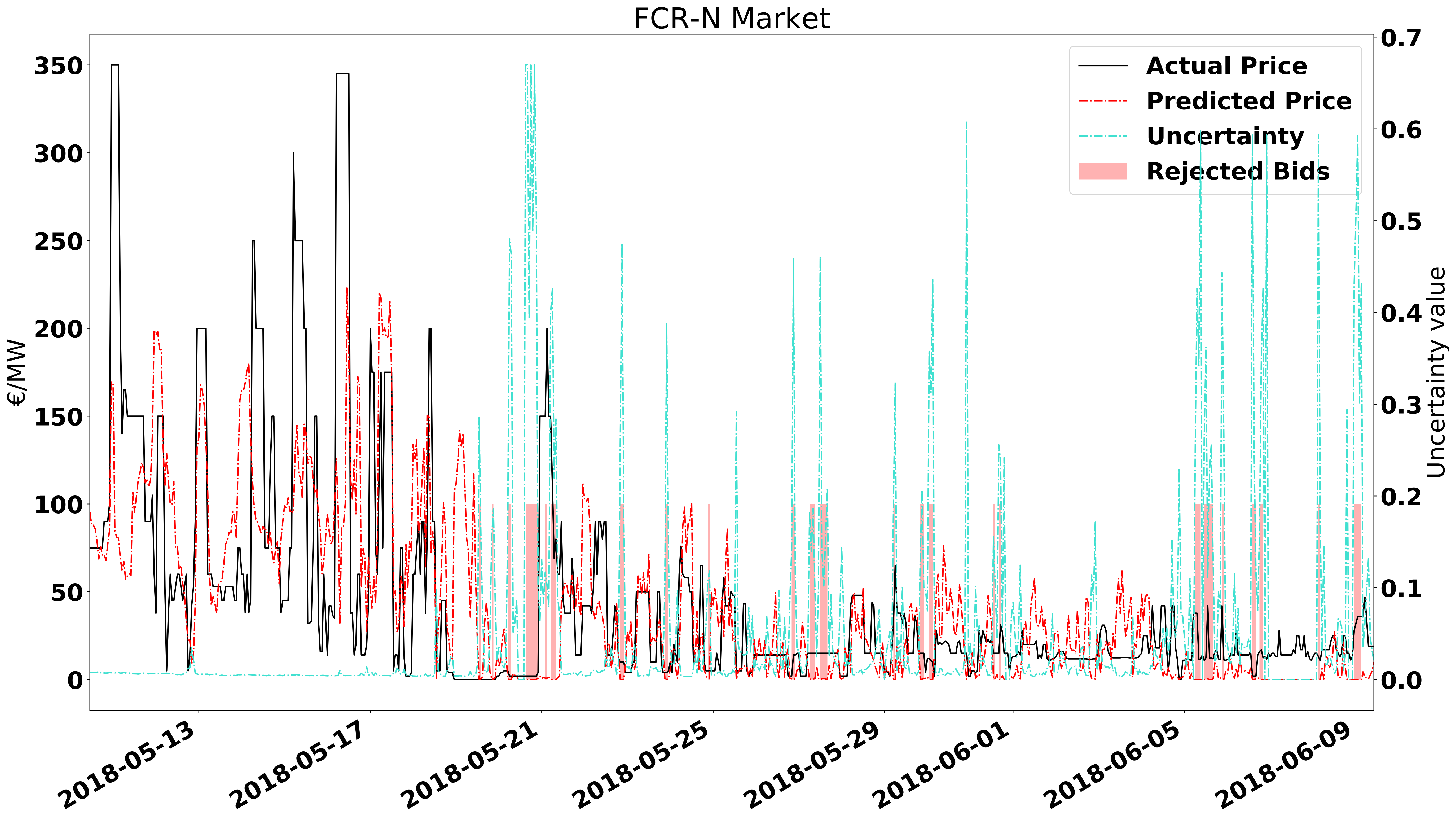} 
  \caption{FCR-N market}
  \label{fig:fcr_n_price}
\end{subfigure}%
\newline
\begin{subfigure}{.5\textwidth}
  \centering
  \includegraphics[scale=0.16]{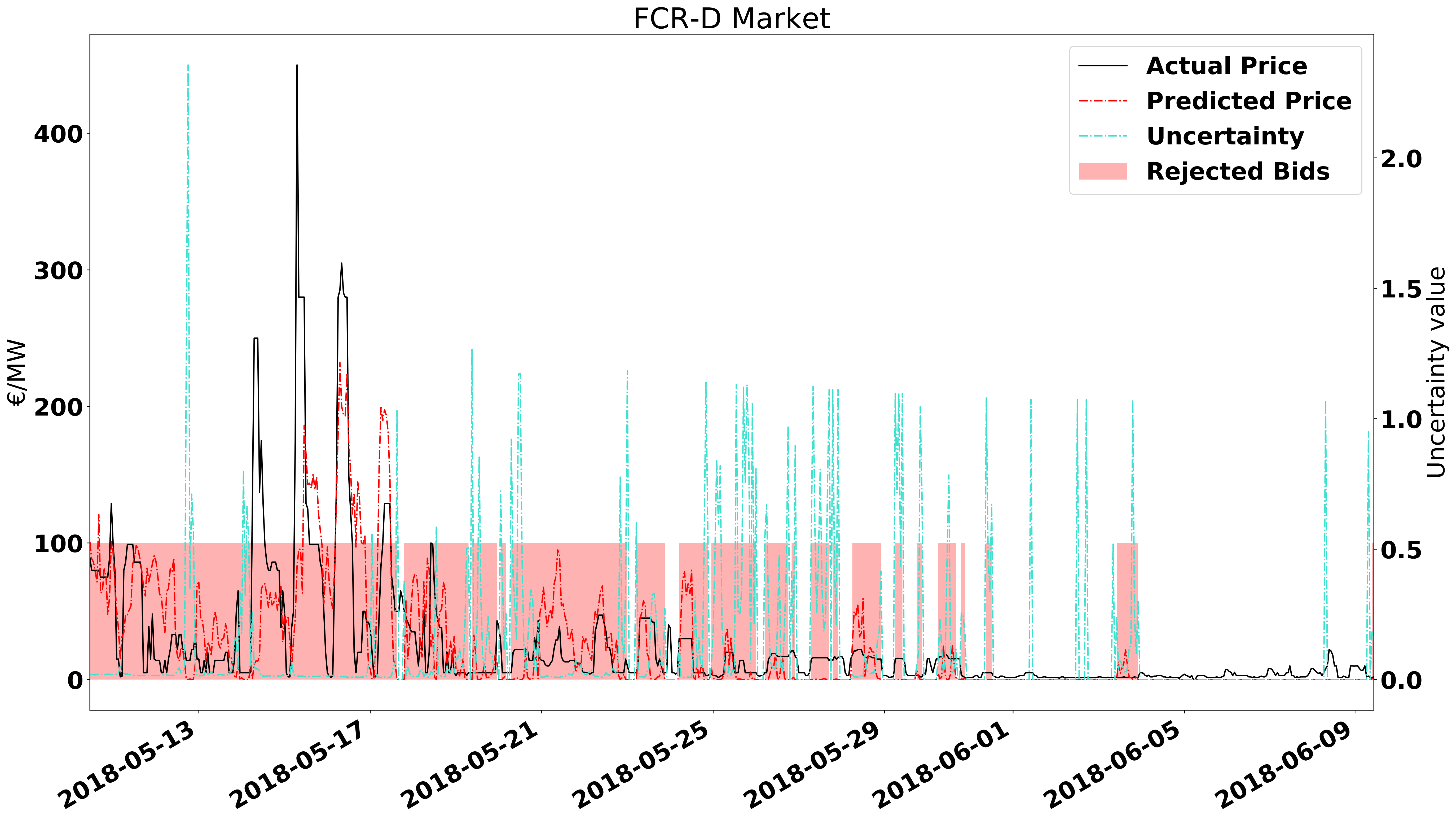}
  \caption{FCR-D market}
  \label{fig:fcr_d_price}
\end{subfigure}
\newline
\begin{subfigure}{.5\textwidth}
  \centering
  \includegraphics[scale=.16]{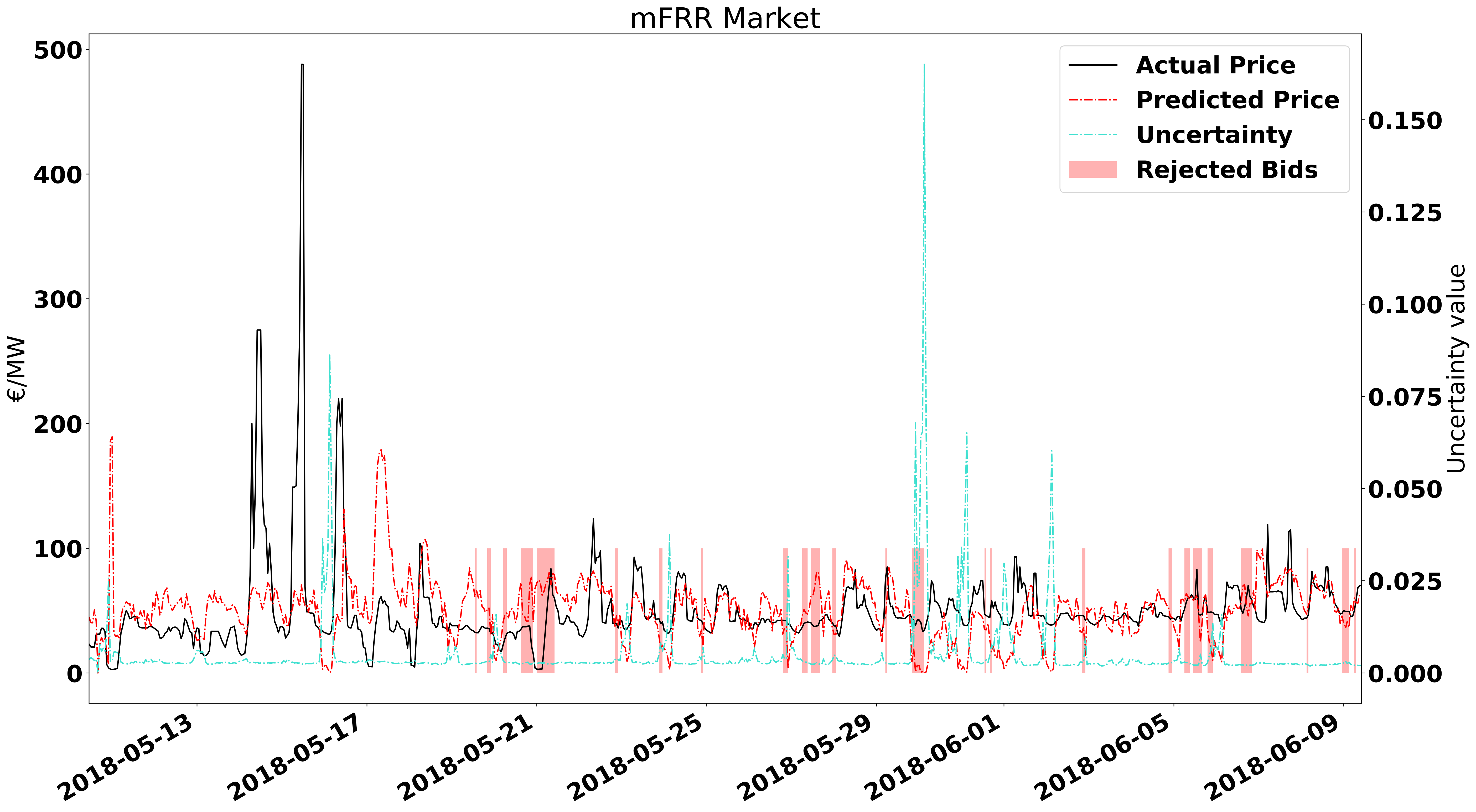}
  \caption{mFRR market}
  \label{fig:mfrr_price}
\end{subfigure}
\caption{Price prediction with Uncertainty : Actual price against predicted price and predicted uncertainty value, rejected bids are identified using the uncertainty threshold value in Table 5}
\label{fig:price}
\end{figure}
Uncertainty accuracy is quantified using the proposed uncertainty accuracy (UA) metric and uncertainty measure from MC-Dropout and GSOM. The highest accuracy generated is 78.88\%. Compared to the other two markets FCR-D market has more uncertain price ranges corresponding to the highest UA value, as shown in Figure 3-b. \par 
\begin{table}[H]
    \centering
    \begin{tabular}{c|c|c}\hline
         FCR-N&FCR-D&mFRR  \\ \hline
         0.01&0.0&0.17 
    \end{tabular}
    \caption{Selected uncertainty threshold values}
\end{table}
Selection of uncertainty thresholds was resolve as a maximizing optimization problem. For the optimization problem, we used uncertainty accuracy (UA) as the maximizing cost function and used hierarchical search method to find the optimum threshold values. Table 5 shows the selected threshold values for each market for the UA value of 78.88\%. 

\subsection{Experiment 2:Bidding Strategy 1 and 2 analysis}
This experiment analyses  bidding strategy 1 and 2 using the predicted price, uncertainty measure and uncertainty threshold values for highest UA value identified in experiment 1. Each strategy selects the highest priced market from the three markets to bid for each epoch. Figure 4 and 5 show the selected market vs the actual price for the three markets. The selected market is depicted as a red dot in actual price graphs.\par
\begin{figure}[H]
	\centering
		\includegraphics[scale=.16]{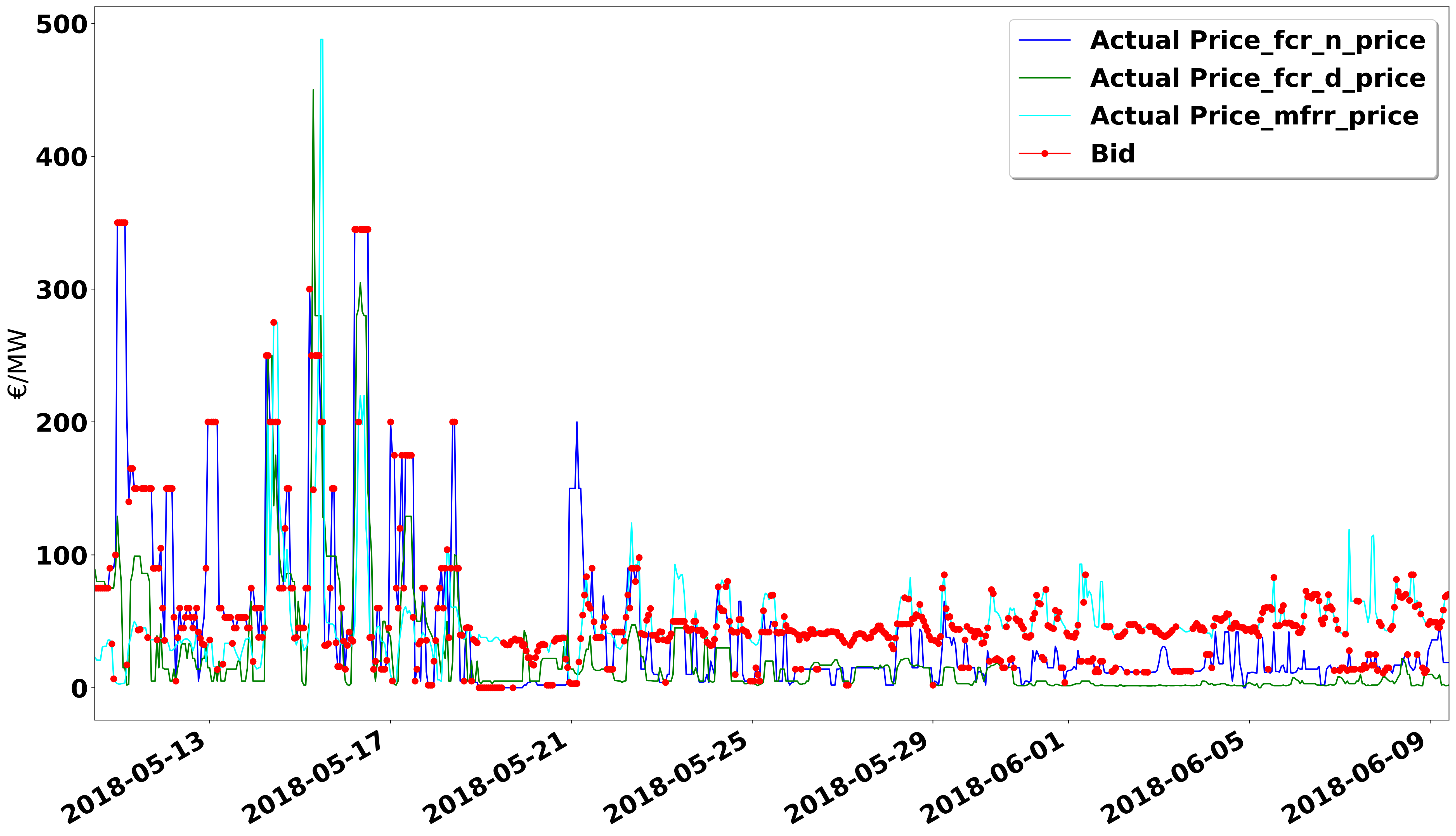}
\caption{Bidding Strategy 1 : Actual prices of FCR-N,FCR-D and mFRR markets for the date range 2018-05-10 to 2018-06-8 with identified bids in red dots}
\end{figure}
\begin{figure}[H]
	\centering
		\includegraphics[scale=.16]{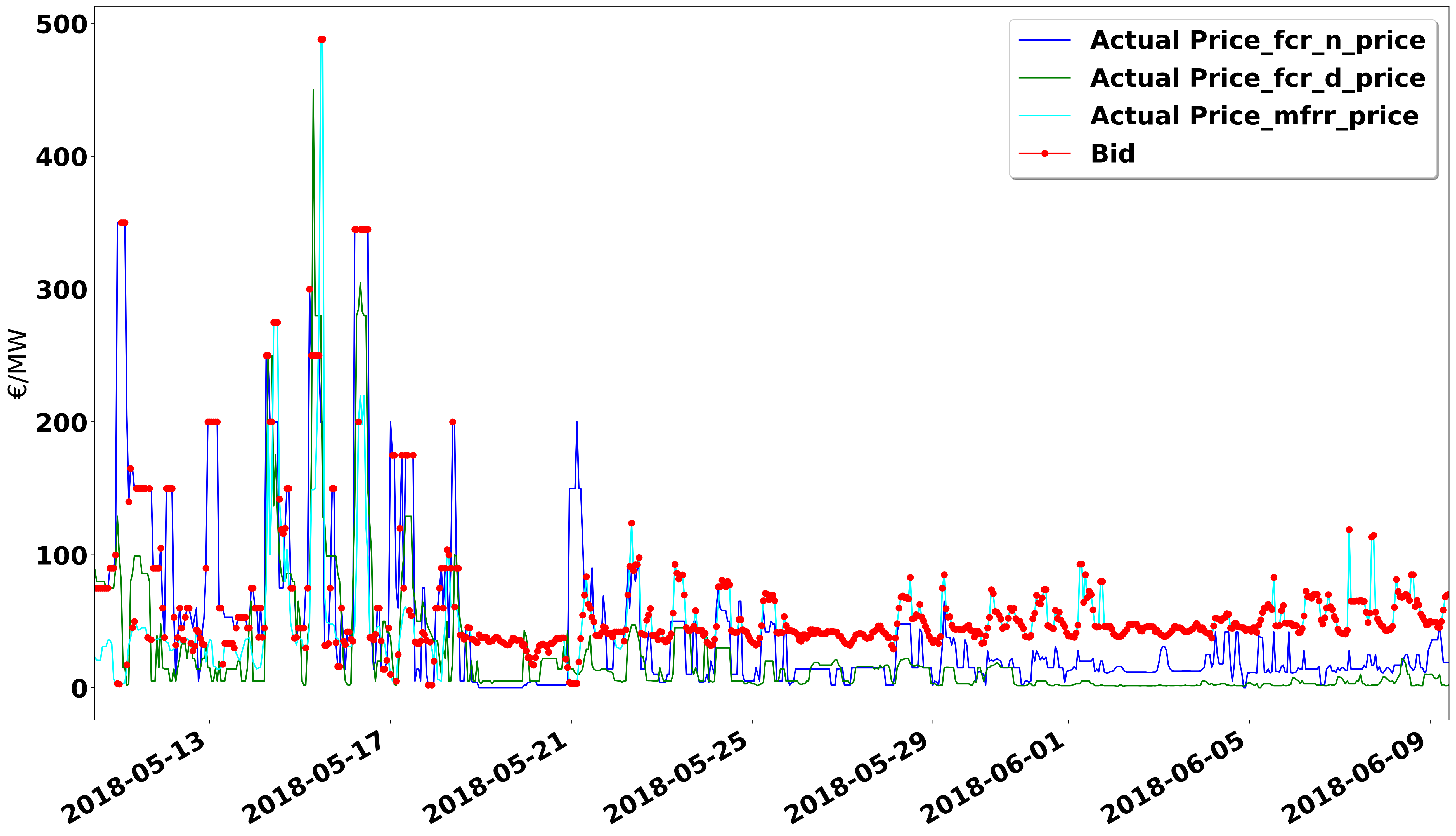}
\caption{Bidding Strategy 2 : Actual prices of FCR-N,FCR-D and mFRR markets for the date range 2018-05-10 to 2018-06-8 with identified bids in red dots}
\end{figure}
According to Figure 4 and 5, bidding strategy 2 has selected most of the actual highest priced markets for each epoch compared to the bidding strategy 1 which can be confirmed using the two dotted lines in red: high and low price points selected in Figure 4.

\subsubsection{Bidding Strategy 1 and 2 comparison}

\begin{figure}[H]
\begin{subfigure}{.25\textwidth}
  \centering
  \includegraphics[width=1.2\linewidth]{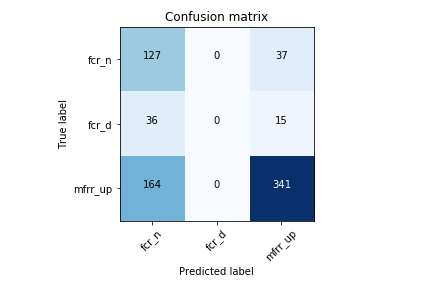}
  \caption{Bidding Strategy 1}
  \label{fig:b_1_confusion}
\end{subfigure}%
\begin{subfigure}{.25\textwidth}
  \centering
  \includegraphics[width=1.2\linewidth]{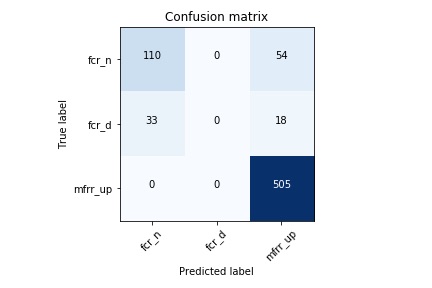}
  \caption{Bidding Strategy 2}
  \label{fig:b_2_confusion}
\end{subfigure}
\caption{Confusion Matrix of Bidding Strategy 1 and 2}
\label{fig:confusion}
\end{figure}

\begin{table}[H]
    \centering
    \begin{tabular}{c|c}\hline
         Method&Accuracy(\%)  \\\hline
         Without bidding optimization&75.41 \\\hline
         Bidding strategy 1&65 \\\hline 
         Bidding strategy 2&85.55 \\\hline 
    \end{tabular}
    \caption{Bidding Strategy 1 and 2 comparison with accuracy}
\end{table}
Experiment 2 is evaluated as a binary classification problem, for each epoch if the max selected market from bidding strategy is equal to the actual max priced market it is considered as a correct classification. Figure 6 shows the confusion matrix for the two bidding strategies for predicted and actual values for each market. According to the confusion matrix, both bidding strategy 1 and 2 has the highest accuracy for the mFRR market and lowest for the FCR-D.\par
Table 6 shows the classification accuracy of the two bidding strategies against without bidding strategy with price prediction. Bidding strategy 2 outperforms the other two methods.\par

\begin{table}[H]
    \centering
    \begin{tabular}{c|c|c}\hline
         &Bidding strategy 1&Bidding strategy 2  \\\hline
         With uncertainty&41315.92&45796.95 \\\hline
         Without uncertainty&31647.73&46089.27 \\\hline
    \end{tabular}
    \caption{Bidding Strategy 1 and 2 comparison with uncertainty}
\end{table}
Further, experiment 2 is evaluated with and without uncertainty. For each method, revenue is calculated assuming the 10MW constant supply for each epoch. Total revenue of each method is compared in Table 7. According to the results, bidding strategy 1 has a significant increase in revenue with uncertainty measure, but bidding strategy 2 has a slight decrease with uncertainty measure compared to without uncertainty.
With bidding strategy 1, the hour-ahead market mFRR is traded on only when there are high uncertainty values for both day-ahead markets FCR-N and FCR-D. Due to frequent high prices on mFRR, the use of uncertainty with bidding strategy 1 has resulted in increased revenue. This phenomenon does not occur with bidding strategy 2, since it refrains from bidding on day-ahead markets if a higher price has been forecasted on the hour-ahead market. The difference between bidding strategies 1 and 2 is that strategy 1 relies more on AI whereas additional knowledge of multi-stage markets has been incorporated to bidding strategy 2. According to Table 7, the uncertainty metric is useful to overcome the lack of knowledge on multi-stage markets in the case of bidding strategy 1.

\subsection{Experiment 3:Bidding strategy with reschedulable loads}

This experiment evaluates bidding strategy 3 for reschedulable loads. This experiment is also based on the price prediction, uncertainty measure and uncertainty thresholds acquired in experiment 1 in 7.1 Reschedulable loads are evaluated according to the EV vehicle scenario mentioned in the Section 4.3 Bidding Strategy 3. 4 Schemes are compared including the novel method based on predicted price and uncertainty value.\par

\begin{figure*}[hbt]
\begin{subfigure}{0.5\textwidth}
  \centering
		\includegraphics[width=0.9\linewidth]{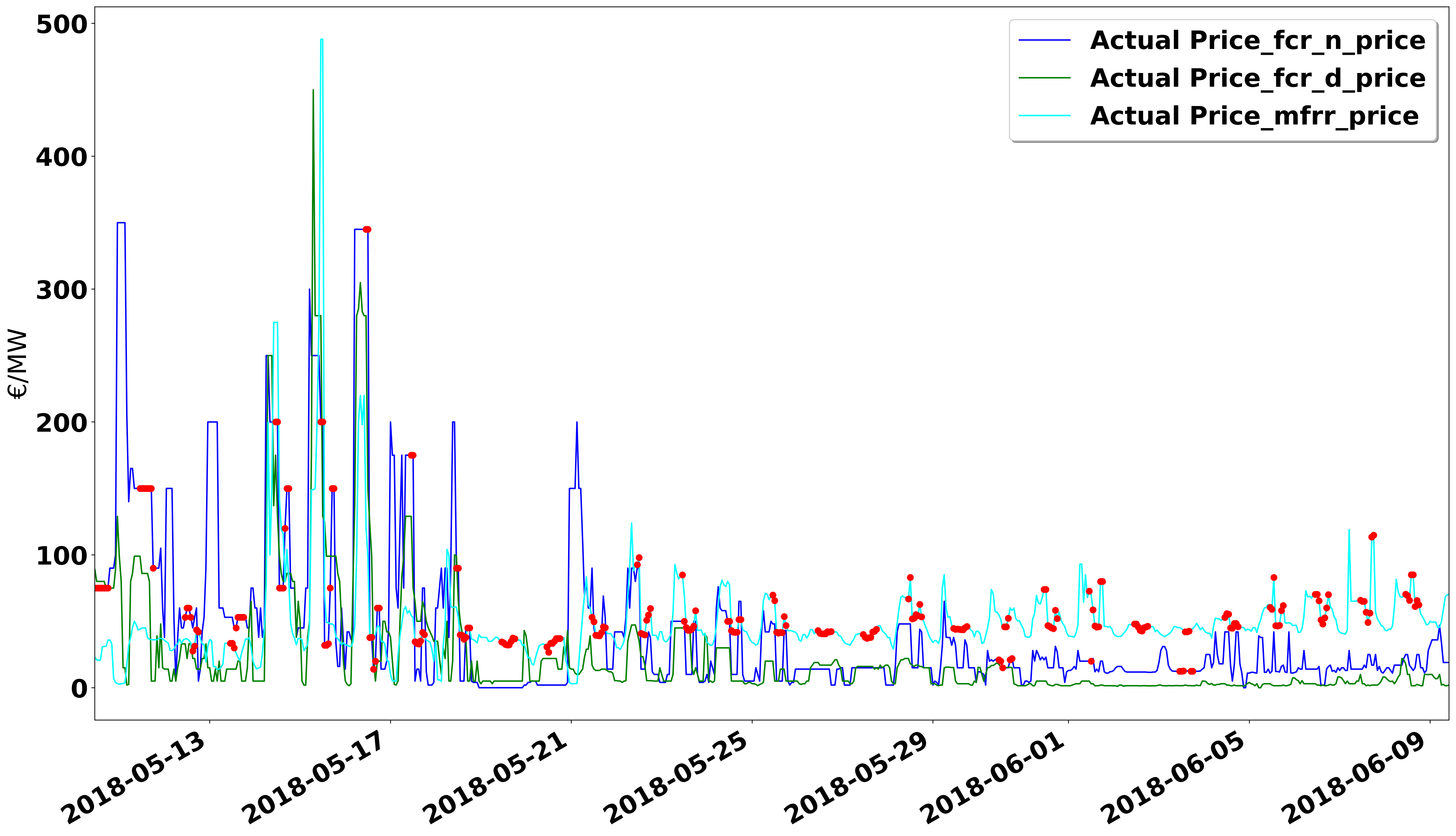}
  \caption{Scheme 1}
  \label{fig:scheme1}
\end{subfigure}%
        \hfill
\begin{subfigure}{0.5\textwidth}
  \centering
		\includegraphics[width=0.9\linewidth]{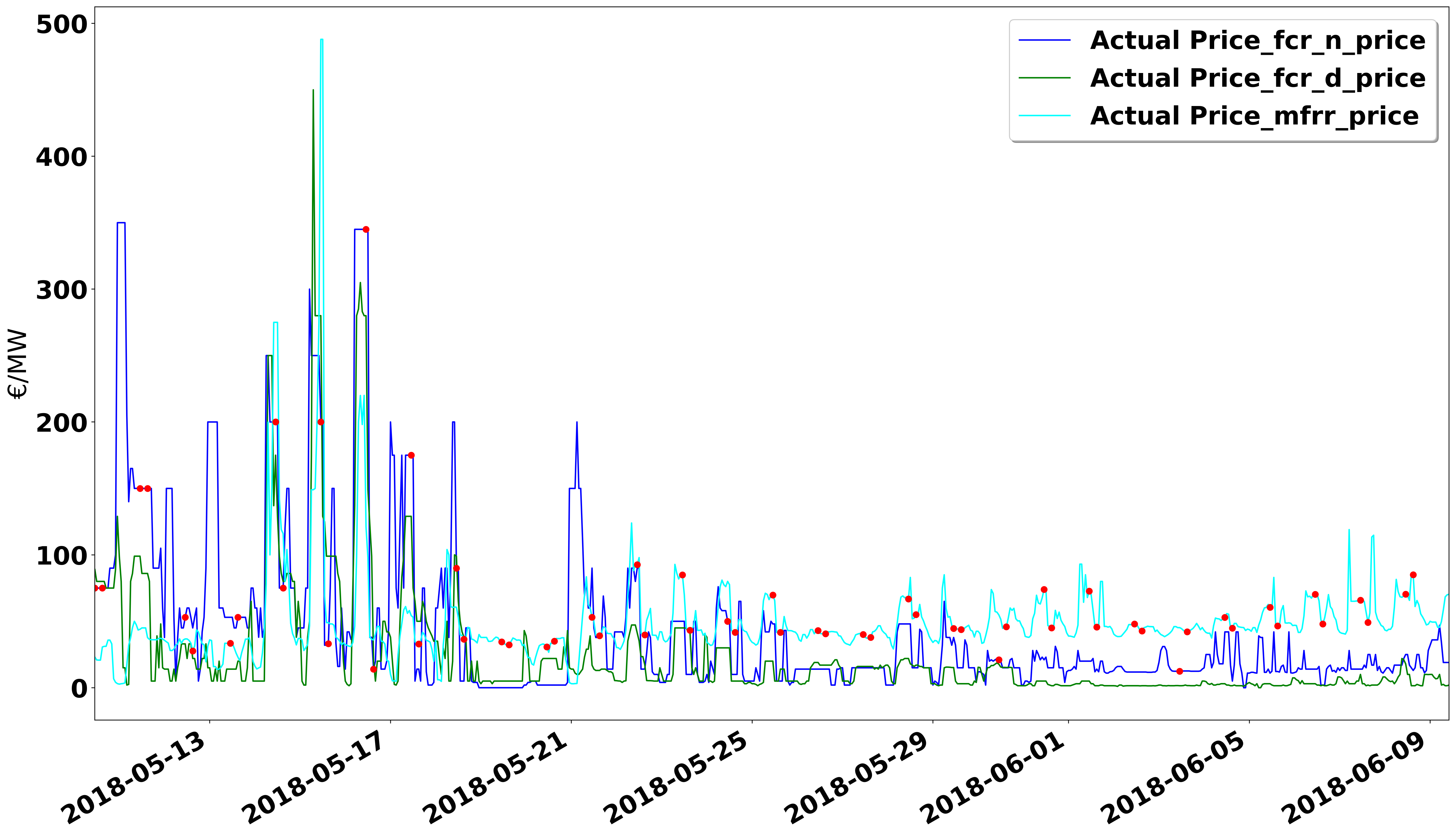}
  \caption{Scheme 2}
  \label{fig:scheme2}
\end{subfigure}%
\hfill
\begin{subfigure}{0.5\textwidth}
  \centering
		\includegraphics[width=0.9\linewidth]{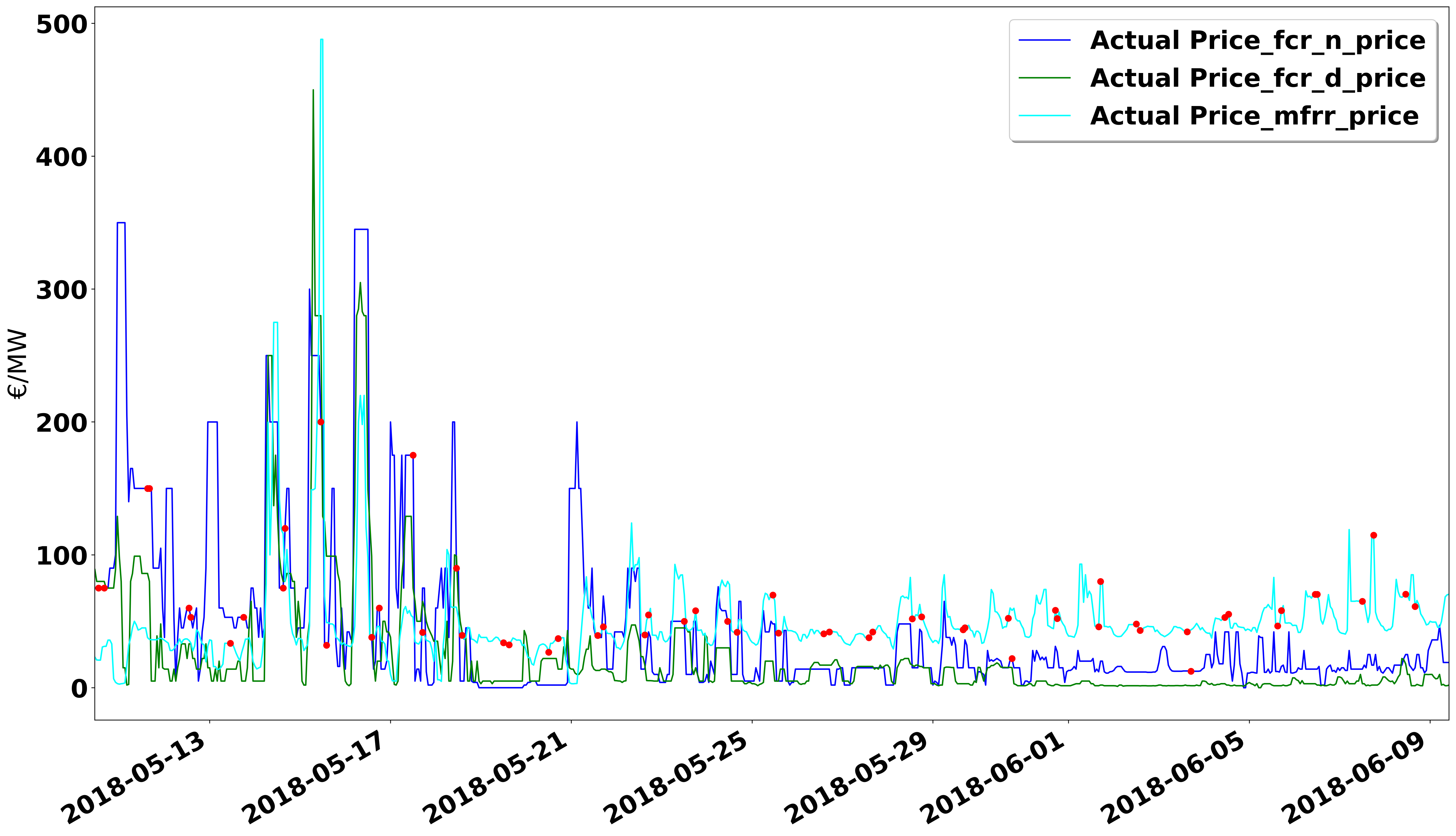}
  \caption{Scheme 3}
  \label{fig:scheme3}
\end{subfigure}
\hfill
\begin{subfigure}{0.5\textwidth}
  \centering
		\includegraphics[width=0.9\linewidth]{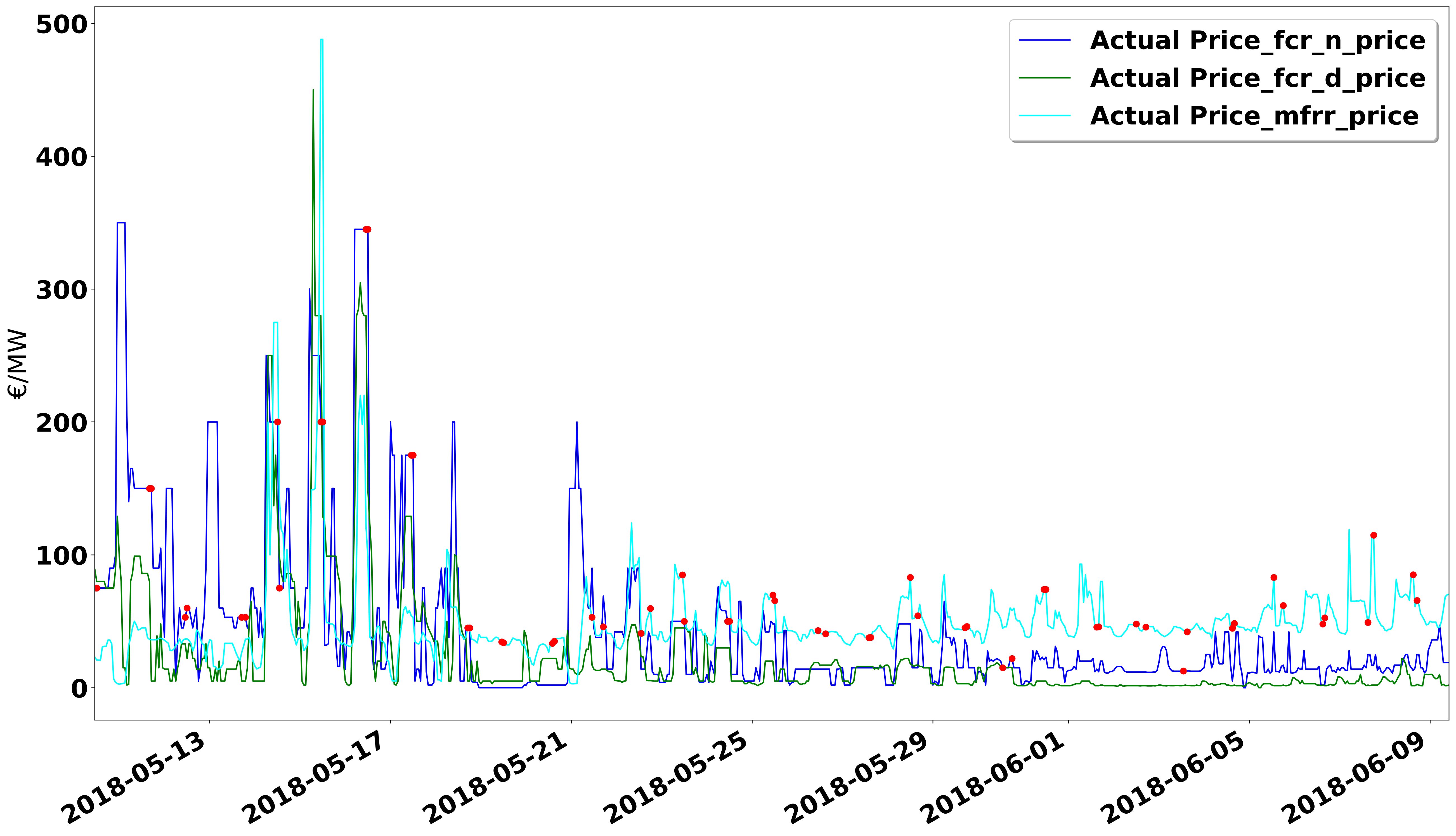}
  \caption{Scheme 4}
  \label{fig:scheme4}
\end{subfigure}
\caption{Bidding schedule comparison of reschedulable loads : Actual prices of FCR-N,FCR-D and mFRR markets for the date range 2018-05-10 to 2018-06-8 with identified bids in red dots }
\label{fig:reschedulable_1_4}
\end{figure*}
Figure 7 shows the results for each scheme of this bidding strategy. The red dots indicate the bidding market for each selected epoch with amount $C[e]$. Compared to the scheme 1,2 and 3, scheme 4 capture most of the price peaks in all three markets.\par

\subsubsection{Bidding strategy with 
reschedulable loads comparison}
Experiment 3 is evaluated by comparing the revenue generated from each scheme. Figure 8 and 9 depict the comparison of the scheme 1 to 4 revenue.\par
\begin{figure}[H]
	\centering
		\includegraphics[scale=.15]{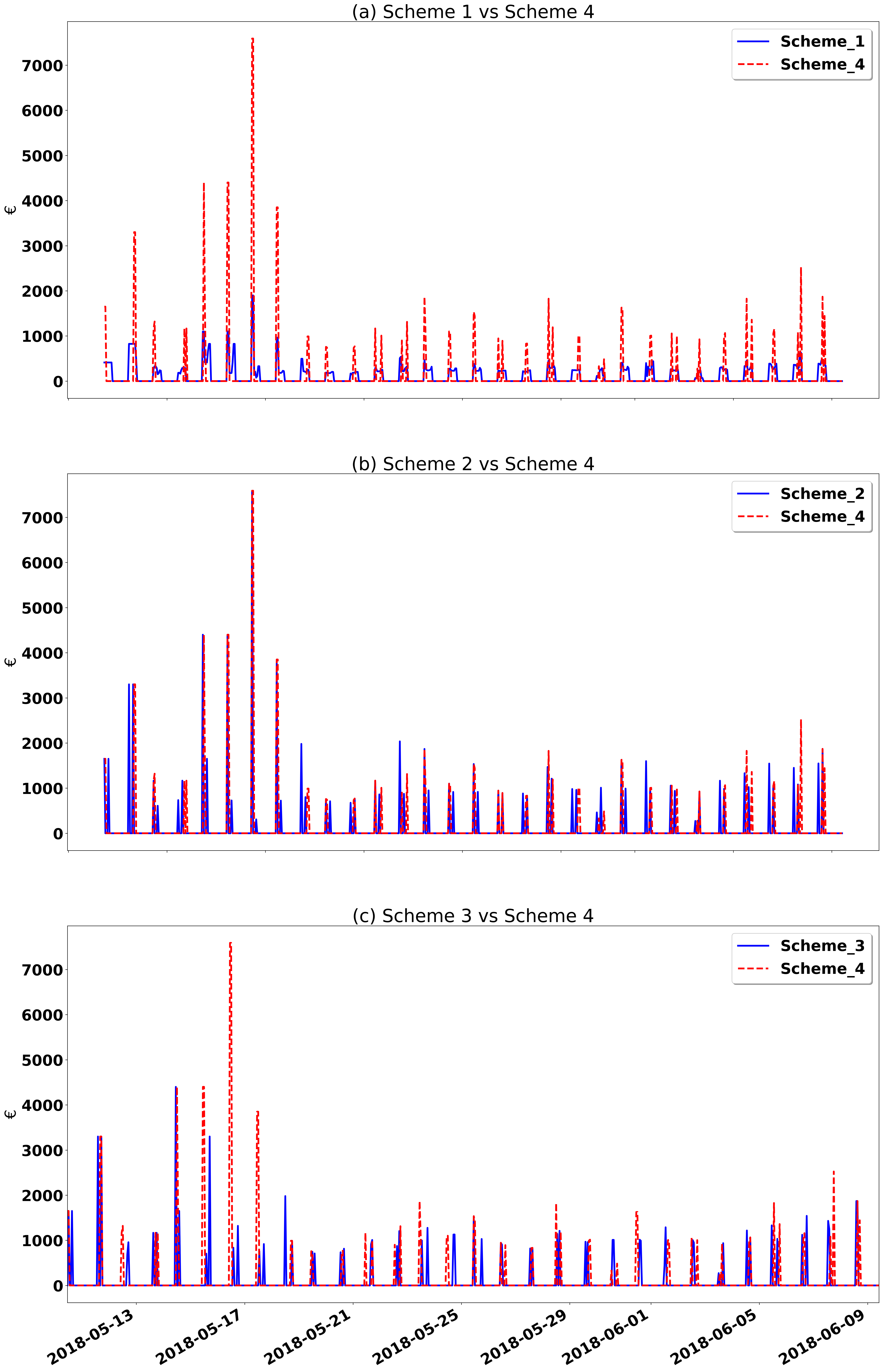}
\caption{Bidding strategy 3 with reschedulable loads profit comparison with individual schemes compared to proposed scheme 4}
\end{figure}

\begin{figure}[H]
	\centering
		\includegraphics[scale=0.16]{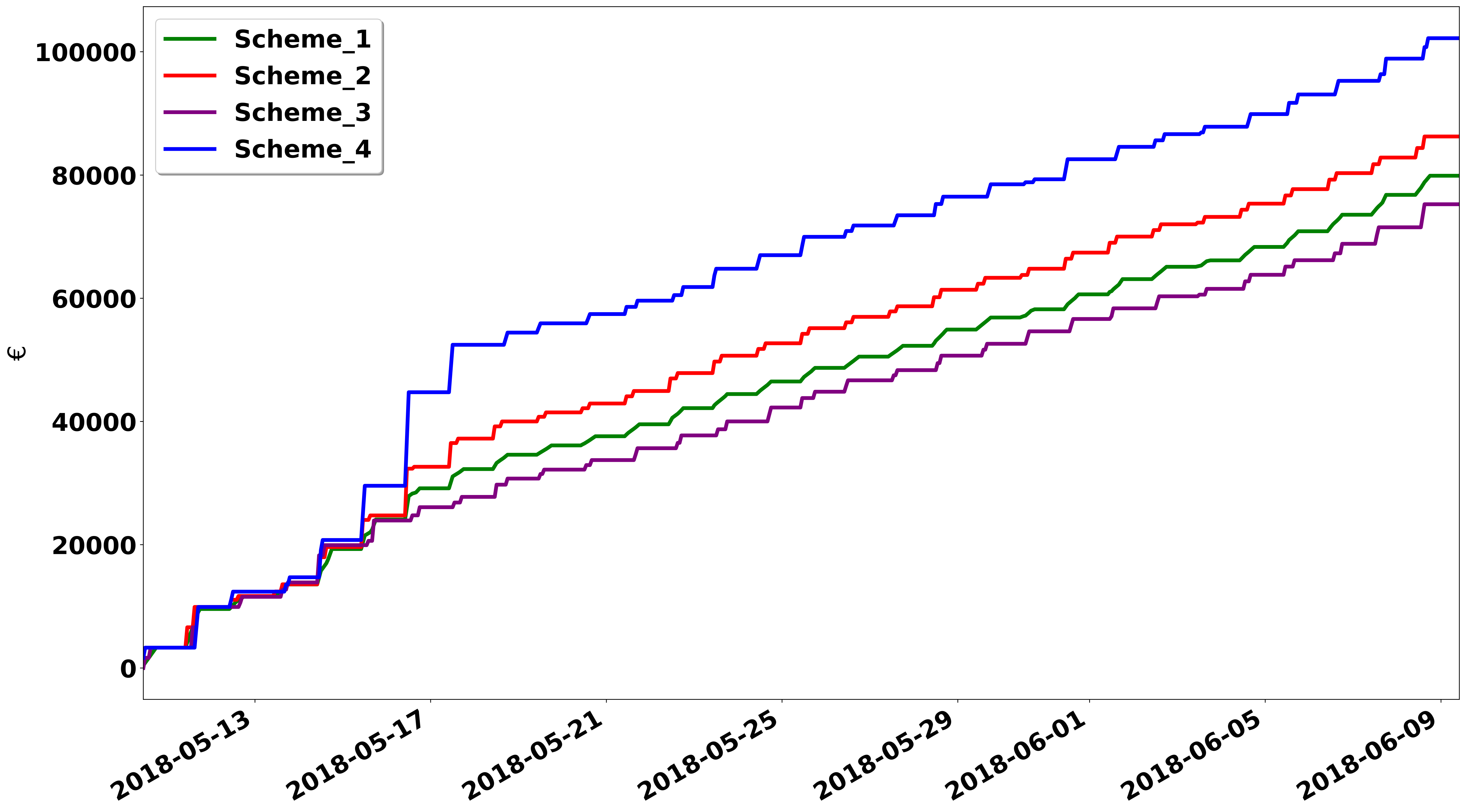}
\caption{Bidding strategy 3 with reschedulable loads cumulative profit comparison, proposed scheme 4 has significant different compared to scheme 1,2 and 3}
\end{figure}
Figure 8 shows the revenue of each epoch of scheme 1,2 and 3 compared to scheme 4. According to the graphs, scheme 4 has significant difference in individual epoch revenue compared to scheme 1 and 3. Scheme 2 has slight changes in epoch times. Figure 9 (cumulative revenue graph) further confirms this, according to this graph at the start all four schemes perform equally but after date 2018-05-17 there was a significant increase in the revenue in scheme 4. Moreover, this clearly shows the advantage of slight changes in time between scheme 3 and 4. The reason for this is evident from Figure 7: scheme 4 is the only scheme that is able to identify and bid all the capacity on high FCR-N price peaks using the AI-based price forecasting. Even though some volatile high price points are missed due to the high uncertainty values in scheme 4, it significantly reduces the risk of bidding on low-confidence epochs, leading to a competitive advantage compared to the other three schemes when combined with price forecasting.

\section{Conclusion}
Bidding optimization on multiple frequency reserve markets is a significant innovation that enables our collective clean energy future. In this paper, we presented the design of a generalised model for frequency reserves markets based on an extensive study of such markets in several countries, followed by the formulation of three bidding strategies based on this generalised model. We implemented these three strategies as an AI-based bidding optimization framework, with data-driven price prediction supplemented with uncertainty metrics as inputs to the bidding optimization process in multiple markets. To the best of our knowledge, this is the first study to model the bidding process on frequency reserves markets, formulate bidding strategies for each variation and implement these using an AI framework for optimal bidding outcomes. The framework and corresponding bidding strategies were empirically evaluated on the real-world case study of the Finnish frequency reserves markets. Out of the three bidding strategies, the third with reschedulable loads significantly outperforms the others. In multi-market bidding optimization when a tertiary market has a higher price, all markets with forecasted price should be considered instead of sequential consideration of day-ahead markets followed by hour-ahead markets - this conclusion is supported by the significantly higher revenues obtained by bidding strategy 2 opposed to bidding strategy 1. In summary, these results confirm the effectiveness of the proposed framework for optimized bidding on multiple frequency reserve markets.   
\bibliography{mybibfile}

\appendix
\section{Bidding Strategies}
\begin{figure}[h]
  \centering
  \includegraphics[width=0.8\linewidth]{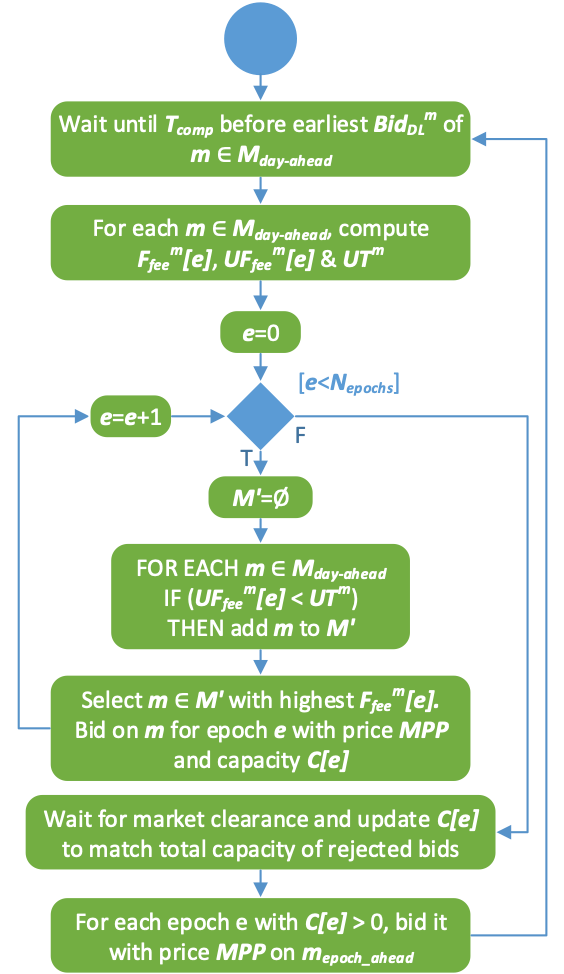}
  \caption{Bidding Strategy 1 UML diagram}
  \label{fig:b_1_uml}
\end{figure}
\begin{figure}[h]
  \centering
  \includegraphics[width=0.8\linewidth]{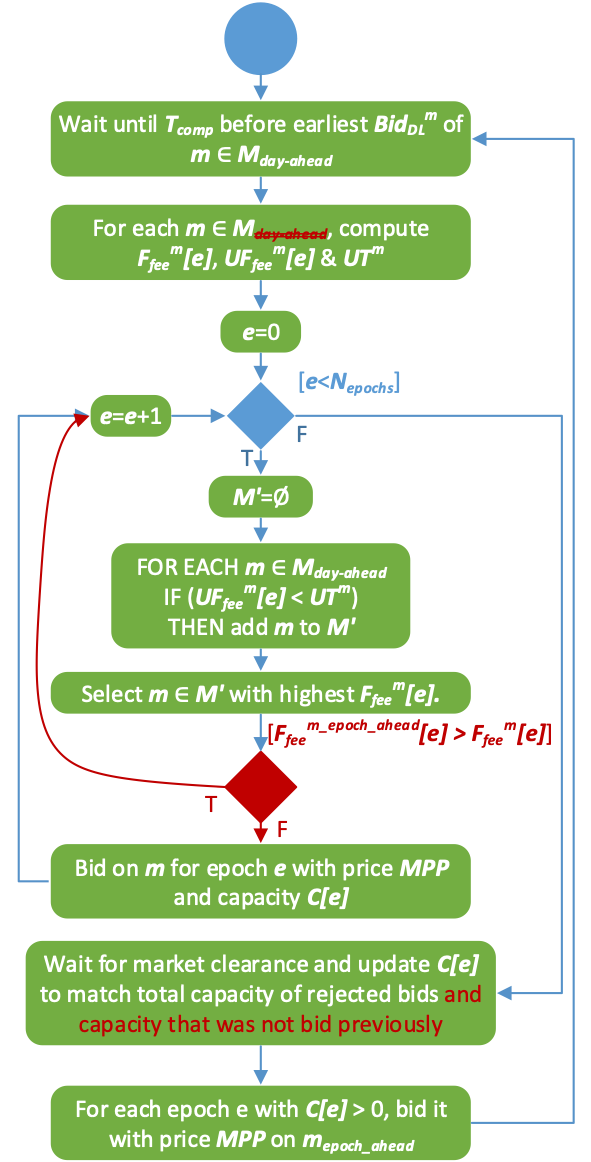}
  \caption{Bidding Strategy 2 UML diagram}
  \label{fig:b_2_uml}
\end{figure}
\begin{figure}[h]
  \centering
  \includegraphics[width=0.8\linewidth]{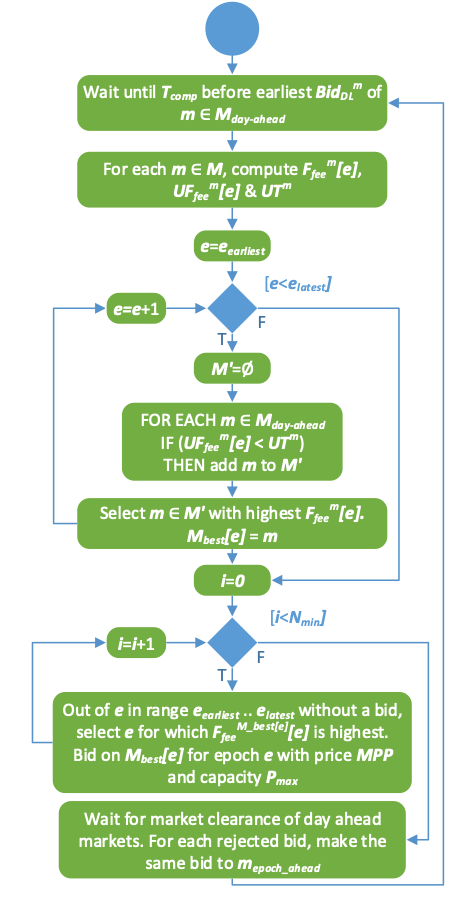}
  \caption{Bidding Strategy 3 UML diagram}
  \label{fig:b_3_uml}
\end{figure}
\end{document}